%% file: PaperTheory8.tex
\documentclass[12pt]{article}

\usepackage[top=30truemm,bottom=30truemm,left=25truemm,right=25truemm]{geometry}
\usepackage{times}
\usepackage{graphicx} 
\usepackage{epsfig} 
\usepackage{algorithm}
\usepackage{algorithmic}
\usepackage{amsmath}
\usepackage{mathrsfs}
\usepackage{amssymb}
\usepackage{threeparttable}
\usepackage[toc,page]{appendix}
\usepackage{url}
\usepackage{fancyhdr}

\usepackage{authblk}

\newcommand{\no}{\nonumber}
\newcommand{\bb}{\boldsymbol}
\providecommand{\keywords}[1]{\textbf{\textit{Key-words:}} #1}

\allowdisplaybreaks

\begin{document}

\author[1]{Tomoki Tokuda \thanks{tomoki.tokuda@oist.jp}} 
\author[1]{Junichiro Yoshimoto}
\author[1]{Yu Shimizu}
\author[2]{Shigeru Toki}
\author[2]{Go Okada} 
\author[2]{Masahiro Takamura}
\author[3]{Tetsuya Yamamoto}
\author[4]{Shinpei Yoshimura}
\author[2]{Yasumasa Okamoto}
\author[2]{Shigeto Yamawaki}
\author[1]{Kenji Doya}


\affil[1]{Okinawa Institute of Science and Technology Graduate University,
             1919-1 Tancha, Okinawa, 904-0495, JAPAN}
\affil[2]{Department of Psychiatry and Neurosciences, Hiroshima University, 
Kasumi1-2-3, Minami-ku, Hiroshima, 734-8851, Hiroshima, JAPAN}
 \affil[3]{Japan Society for the promotion of science}
\affil[4]{Otemon Gakuin University, Nishiai 2-1-15, Ibaraki city, Osaka,  567-8502, JAPAN}

 \pagestyle {fancy}
\rhead{Nonparametric multiple co-clustering method}

\setcounter{Maxaffil}{0}
\renewcommand\Affilfont{\itshape\small}
\date{}
\title{Multiple co-clustering based on nonparametric mixture models with heterogeneous marginal distributions
}

\maketitle

\begin{abstract} 
We propose a novel method for multiple clustering that assumes a co-clustering structure (partitions in both rows and columns of the data matrix) in each view.  The new method is applicable to high-dimensional data. It is based on a nonparametric Bayesian approach in which the number of views and the number of feature-\slash subject clusters are inferred in a data-driven manner. We simultaneously model different distribution families, such as Gaussian, Poisson, and multinomial distributions in each cluster block. This makes our method applicable to datasets consisting of both numerical and categorical variables, which biomedical data typically do. Clustering solutions are based on variational inference with mean field approximation. We apply the proposed method to synthetic and real data, and show that our method outperforms other multiple clustering methods both in recovering true cluster structures and in computation time. Finally, we apply our method to a depression dataset with no true cluster structure available, from which useful inferences are drawn about possible clustering structures of the data.
\end{abstract}


\section{Introduction}
\begin{figure}
\centering
\includegraphics[scale=0.143, trim=50mm 10mm 10mm 10mm]{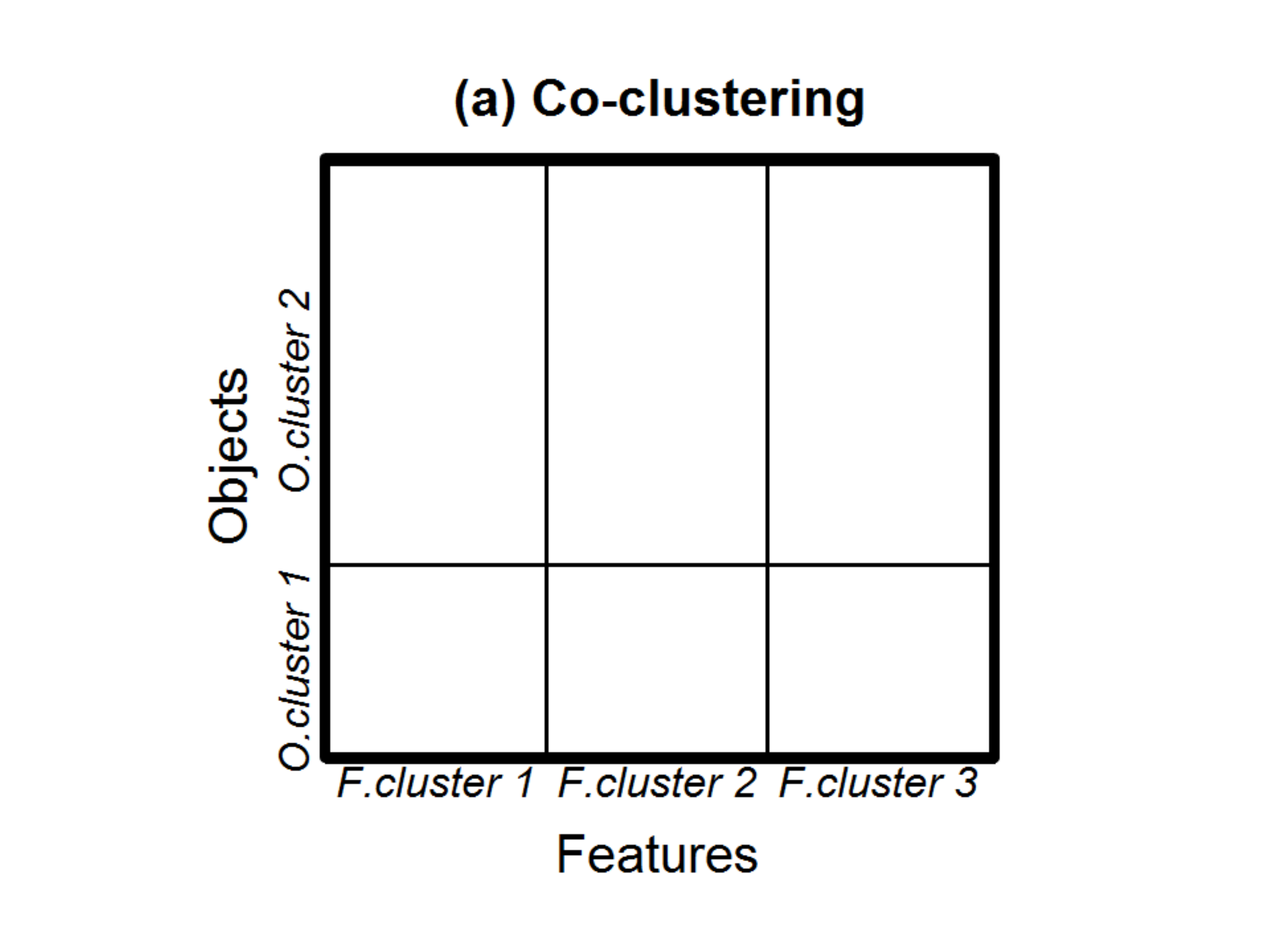} 
\includegraphics[scale=0.143, trim=50mm 10mm 10mm 10mm]{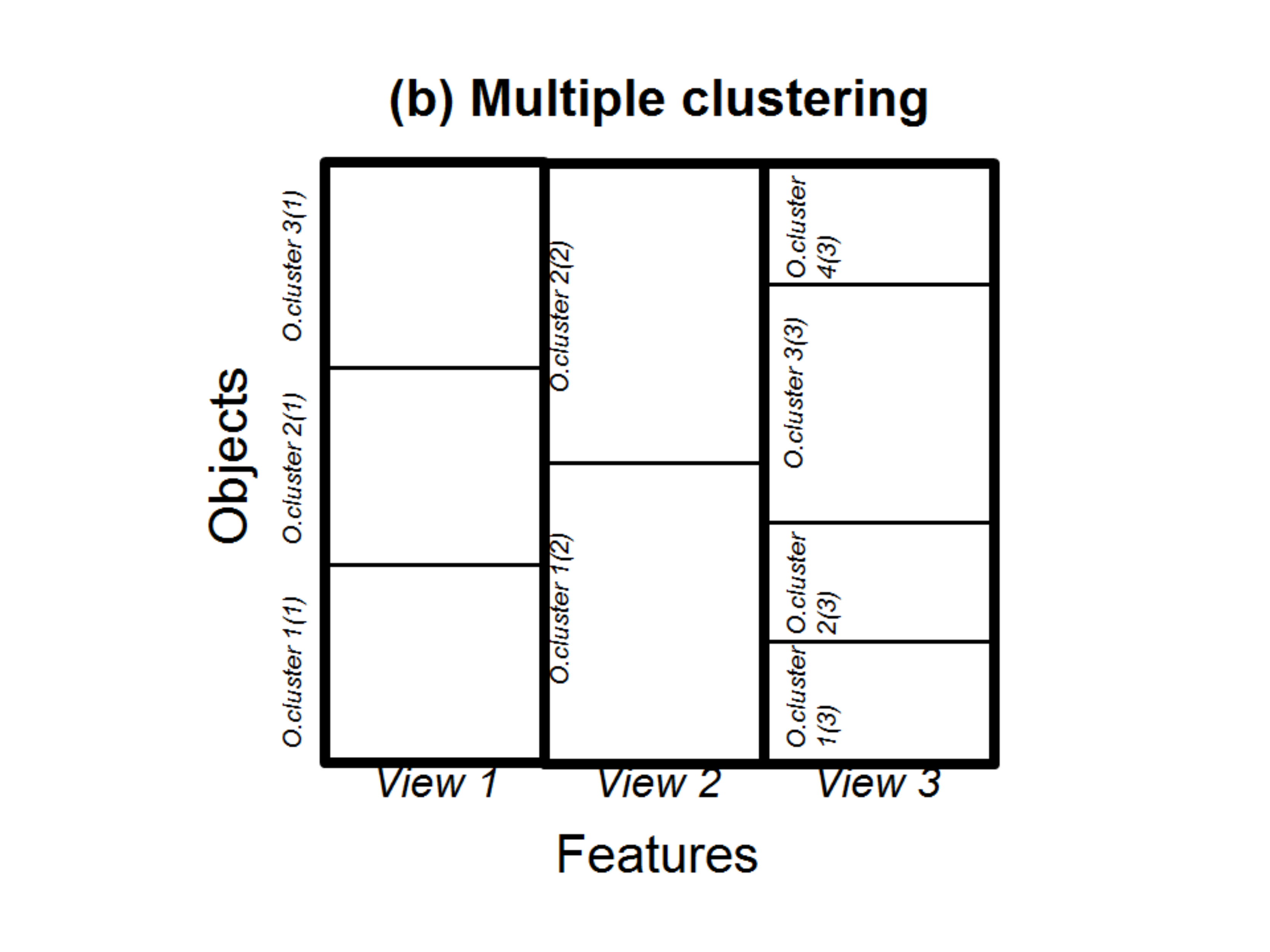} 
\includegraphics[scale=0.143, trim=50mm 10mm 10mm 10mm]{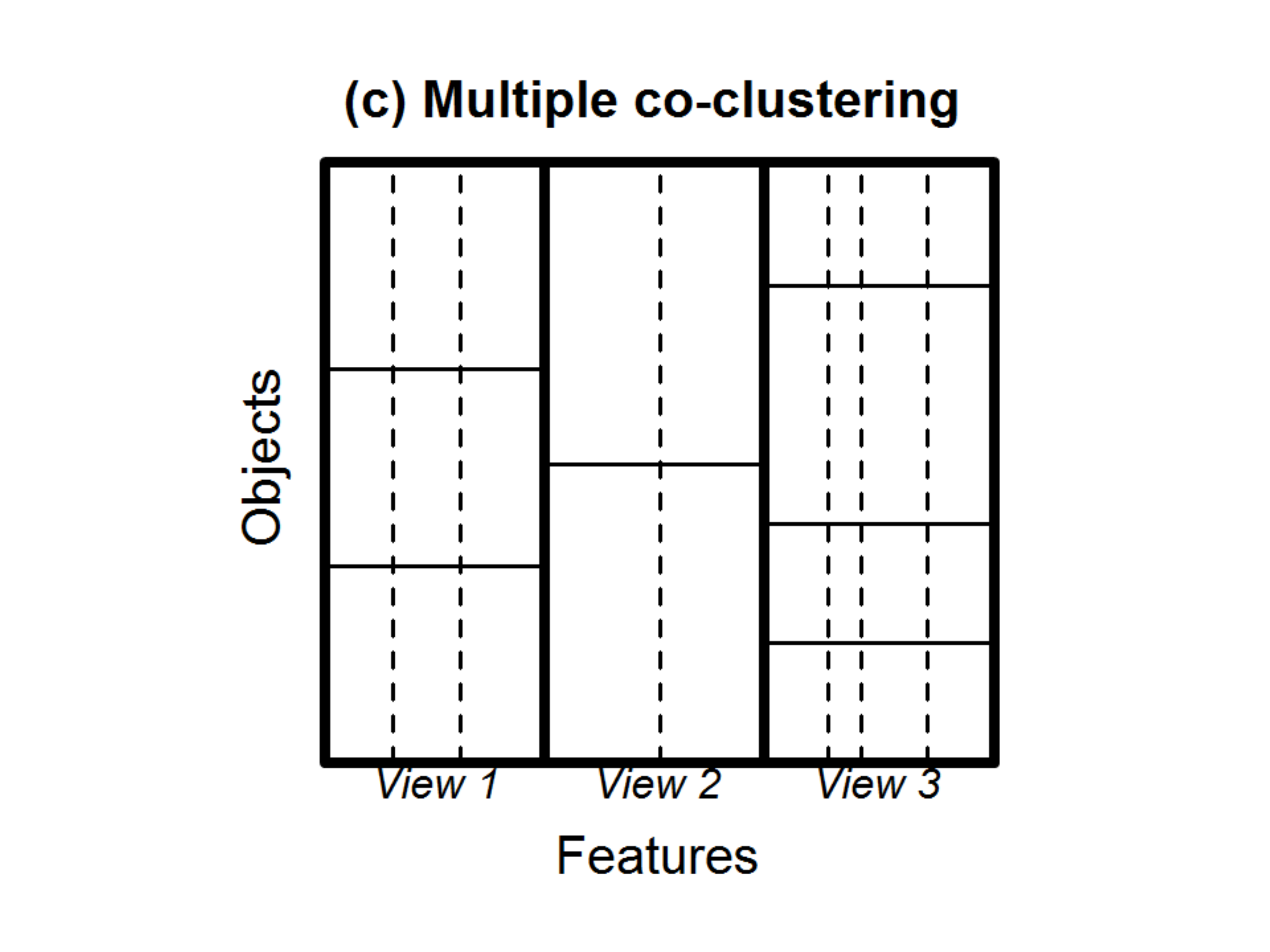}
\includegraphics[scale=0.143, trim=50mm 10mm 10mm 10mm]{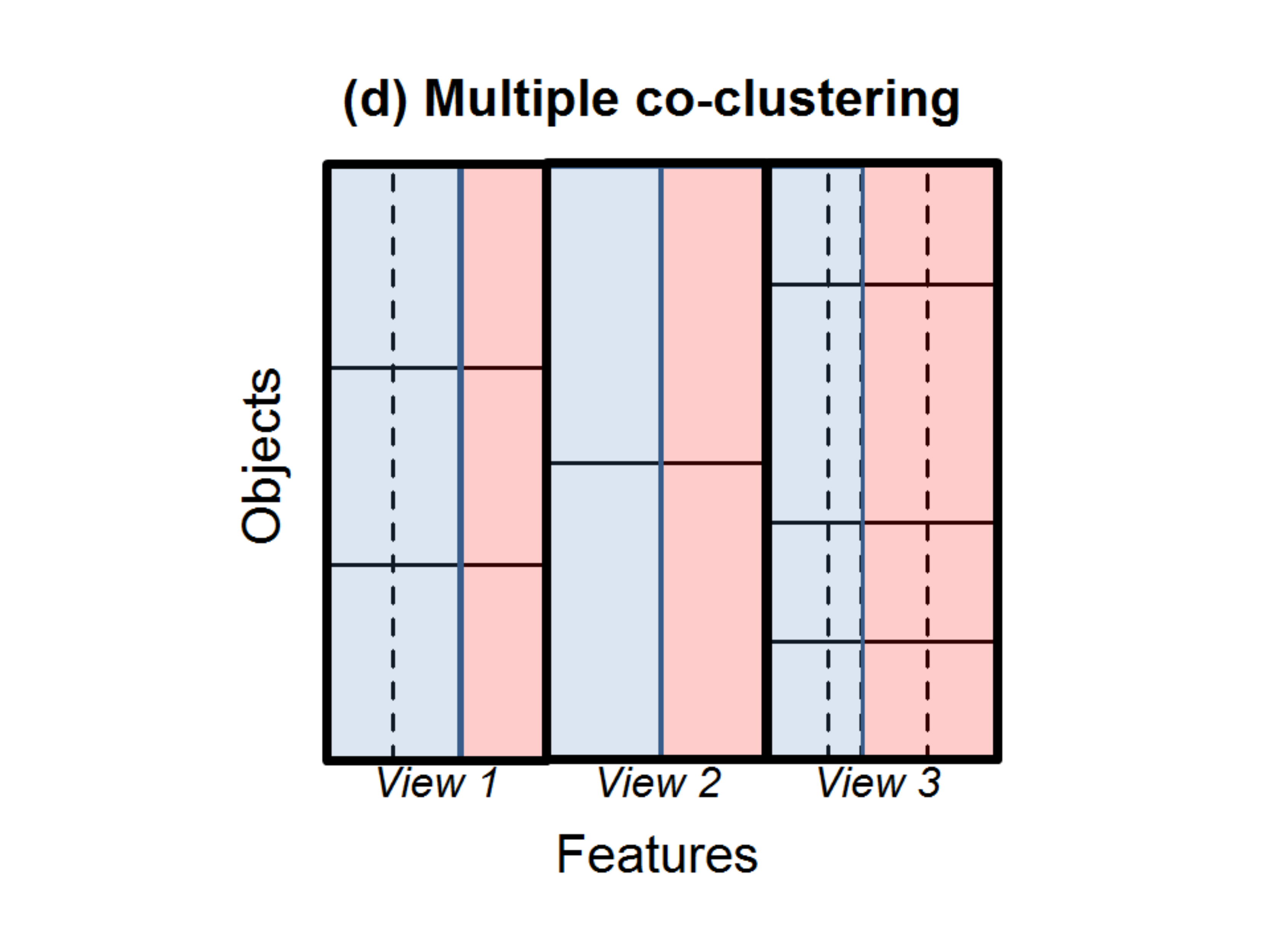} 
\caption{\it  \small Illustration of clustering structures: Panel (a) co-clustering;
(b)  multiple clustering (with full covariance of Gaussian); (c) multiple clustering with a specific structure of co-clustering; 
(d) extension of the model (c) where different distribution families are mixed 
(two distributions families in blue and red). Note that a rectangle surrounded by bold lines corresponds to a single 
co-clustering structure with a single object cluster solution.  
In these panels, features and objects are sorted in the order of view, feature and object cluster indices 
(hence, the order of objects differs among the co-clustering rectangles).
}  
\label{illustration1}
\end{figure}

We consider a clustering problem for a data matrix that consists of objects in rows and features (variables, or attributes) in columns. Clustering objects based on the data matrix is a basic data mining approach, which groups objects with similar patterns of distribution. As an extension of conventional clustering, a co-clustering model has been proposed, which captures not only object cluster structure, but also feature cluster structure (features are grouped based on their distribution patterns, Figure~\ref{illustration1}a). This has the effect of reducing the number of parameters, which enables the model to fit high-dimensional data. Yet, the co-clustering method (as well as conventional clustering methods) often does not work well for high-dimensional data, because such data may have different `views' that characterize multiple clustering solutions \footnote[1]{We also use a terminology of `clustering', meaning the whole set of clusters in a view.}  \cite{muller2012discovering, niu2010multiple}. For instance, in DNA analysis, a set of genes (features) may be related to clustering of subjects for a specific genetic disorder, while another set may be related to clustering of subjects for a different disorder.

Recently, several heuristic methods have been proposed to detect multiple clustering solutions of objects \cite{caruana2006meta, bae2006coala, jain2008simultaneous, dang2010generation}. However, with these multiple clustering methods it is not straightforward to determine the number of views. A more promising approach based on nonparametric mixture models assumes multivariate Gaussian mixture models for each view (Figure~\ref{illustration1}b)
\cite{guan2010variational}. In the latter approach, the full Gaussian model for covariance matrices is considered, and the numbers of views and of object clusters are inferred in a data-driven way via the Dirichlet process. Such a method is quite useful to discover possible multiple cluster solutions when these numbers are not known in advance, which is usually the case. However, its application is limited to low dimensional cases ($p <n$), because in high-dimensional cases, the number of objects to infer posterior distribution for the full covariance matrix of the Gaussian distribution may be insufficient, resulting in overfitting. Moreover, this method also suffers the drawback that features need to belong to the same distribution family.

To address the aforementioned problems, we propose a novel multiple clustering method (referred to hereafter as the multiple co-clustering method). Our model is based on the following extension of the co-clustering model. First, we consider multiple views of co-clustering structure (Figure~\ref{illustration1}c), where a univariate distribution is fitted to each cluster block \cite{Shan2008}. Second, for each cluster block, the proposed method simultaneously deals with an ensemble of several types of distribution families such as Gaussian, Poisson, and multinomial distribution (Figure~\ref{illustration1}d). Obviously, the first extension enables our model to fit high-dimensional data, while the second enables it to fit data that include different types of features (numerical and categorical).

One may consider a multiple clustering model by simply fitting a univariate distribution to each view (hereafter, we call it the `restricted multiple clustering method'). However, such an approach has the drawback that it may replicate similar object cluster solutions for different views. For instance, features that discriminate among object clusters in the same manner would be allocated to different views, if these are negatively correlated or if they have different scales (hence, redundant views). As a consequence, it would not only complicate interpretation, but would also lose discriminative power relative to features. In the present paper, we retain this method for performance comparisons with our method.

\section{Model}
As in \cite{guan2010variational}, our model is based on nonparametric mixture models using the Dirichlet process
\cite{ferguson1973bayesian, escobar1995bayesian}. However, unlike the conventional Dirichlet process, we employ a hierarchical structure, because in our model, the allocation of features is determined in two steps: the first allocation to a view, and the second to a feature cluster in that view. Moreover, we allow for mixing of several types of features, such as mixtures of Gaussian, Poisson, and categorical/multinomial distributions. Note that in this paper, we assume that types of features are pre-specified by the user, and do not draw inferences about them from data. In the following section, we formulate our method to capture these two aspects. To estimate model parameters, we rely on a variational Bayes EM (Expectation Maximization) algorithm, which provides (iterative) updating equations of relevant parameters. In general, determining whether these updating equations may be expressed in closed form is a subtle problem. However, this is the case in our model, which provides an efficient algorithm to estimate views and feature-/object cluster solutions.

\subsection{Multiple clustering model} \label{hyperpara}
We assume that a data matrix $\bb{X}$ consists of $M$ types of distribution families that are known in advance. 
We decompose $\bb{X}=\{\bb{X}^{(1)},\ldots, \bb{X}^{(m)}, \ldots, \bb{X}^{(M)}\}$ with data size $n \times d^{(m)}$
for $\bb{X}^{(m)}$, 
where $m$ is an indicator for a distribution family ($m=1, \ldots, M$). 
Further, we denote the number of views as $V$ (common to all distribution families), the number of feature clusters $G_v^{(m)}$
for view $v$ and distribution family $m$, and the number of object clusters $K_v$ for view $v$ (common to all distribution families). Moreover, for simplicity of notation, we use $G^{(m)}=\max_{v} G_v^{(m)}$ and $K=\max_v K_v$ to denote the number of features and the number of clusters, allowing for empty clusters. 

With this notation, for i.i.d. $d^{(m)}$-dimensional random vectors $\boldsymbol{X}_1^{(m)}, \ldots, \bb{X}_n^{(m)}$ for distribution family $m$, we consider a $d^{(m)}\times V \times G^{(m)}$ feature-partition tensor (3rd-order) $\boldsymbol{Y}^{(m)}$ in which $Y_{j, v, g}^{(m)}=1$ if feature $j$ of distribution family $m$ belongs to feature cluster $g$ in view $v$ (0 otherwise). Combining this for different distribution families, we let $\bb{Y}=\{ \bb{Y}^{(m)}\}_m$.
Similarly, we consider a $n \times V \times K$ object-partition (3rd-order) tensor $\boldsymbol{Z}$ 
 in which $Z_{i, v, k}=1$ if object $i$ belongs to object cluster $k$ in view $v$. Note that feature $j$ belongs to 
one of the views (i.e., $\sum_{v, g}Y_{j, v, g}^{(m)}=1$) 
while object $i$ belongs to each view
(i.e., $\sum_kZ_{i, v, k}^{(m)}=1$).
 Further, $\bb{Z}$ is common to all distribution families, which implies that our model estimates 
subject cluster solutions using information on all distribution families. 

For a prior generative model of $\bb{Y}$, 
we consider a hierarchical structure of views and feature clusters: views are first generated, followed by generation of feature clusters. Thus, features are partitioned in terms of pairs of view and feature cluster memberships, which implies that the allocation of feature is jointly determined by its view and feature cluster.
On the other hand, objects are partitioned into object clusters in each view, hence, 
we consider just a single structure of object clusters for $\bb{Z}$. 
We assume that these generative models are all based on a stick-breaking process as follows.

\subsubsection*{Generative model for feature clusters $\bb{Y}$} 
We let $\bb{Y}_{j\cdot \cdot}^{(m)}$  denote a view/feature cluster membership vector for feature $j$ of distribution family $m$, which is generated by a hierarchical stick-breaking process:
\begin{eqnarray*}
w_{v} &\sim& \mbox{Beta}(\cdot|1, \alpha_1),~v=1, 2, \ldots \\
\pi_{v}&=&w_{v}\prod_{t=1}^{v-1}(1-w_{t}), \\ 
{w'}_{g, v}^{(m)} &\sim& \mbox{Beta}(\cdot|1, \alpha_2),~g=1,2,\ldots, m=1, \ldots, M \\
{\pi '}_{g, v}^{(m)}&=&{w'}_{g,v}^{(m)}\prod_{t=1}^{g-1}(1-{w'}_{t, v}^{(m)}),  \\
 \tau_{g, v}^{(m)} &= &\pi_{v}{{\pi} '}_{g, v}^{(m)}\\
\bb{Y}_{j \cdot \cdot }^{(m)}&\sim &\mbox{Mul}(\cdot|\boldsymbol{\tau}^{(m)}),
\end{eqnarray*}
where $\boldsymbol{\tau}^{(m)}$ denotes a $1 \times GV$ vector $(\tau_{1,1}^{(m)}, \ldots, \tau_{G, V}^{(m)})^T$
(the superscript $T$ denotes matrix transposition); $\mbox{Mul}(\cdot|\boldsymbol{\pi})$ is a multinomial distribution of one sample size with probability parameter $\boldsymbol{\pi}$; 
$\mbox{Beta}(\cdot|a, b)$ is a Beta distribution with prior sample size
$(a, b)$; $\bb{Y}_{j \cdot \cdot}^{(m)}$ is a $1 \times GV$ vector
$(Y_{j, 1, 1}^{(m)}, \ldots, Y_{j, V, G}^{(m)})^T$.
Note that we truncate the number of views with sufficient large $V$ and the number of feature clusters with $G$ \cite{blei2006variational}.
When $Y_{j, v, g}^{(m)}=1$, feature $j$ belongs to feature cluster $g$ at view $v$.
By default, we set the concentration parameters $\alpha_1$ and $\alpha_2$ to one. 

\subsubsection*{Generative model for object clusters $\bb{Z}$}
A subject cluster membership vector of object $i$ in view $v$, denoted as $\bb{Z}_{i, v \cdot}$, is  
generated by
\begin{eqnarray*}
u_{k, v}&\sim& \mbox{Beta}(\cdot|1, \beta),  ~v= 1, 2, \ldots, ~k=1, 2,
 \ldots \\
\eta_{k,v}&=&u_{k, v}\prod_{t=1}^{k-1}(1-u_{t, v}),  \\  
\boldsymbol{Z}_{i,v \cdot} &\sim & \mbox{Mul}(\cdot|\boldsymbol{\eta}_{v}),
\end{eqnarray*}
where $\bb{Z}_{i, v \cdot}$ is a $1 \times K$ (we take $K$ sufficiently large) vector given by $\bb{Z}_{i, v \cdot}=(Z_{i, v, 1}, \ldots, Z_{i, v, K} )^T$.
We set the concentration parameter $\beta$ to one.  

\begin{figure}[!h]
\centering
\includegraphics[scale=0.35]{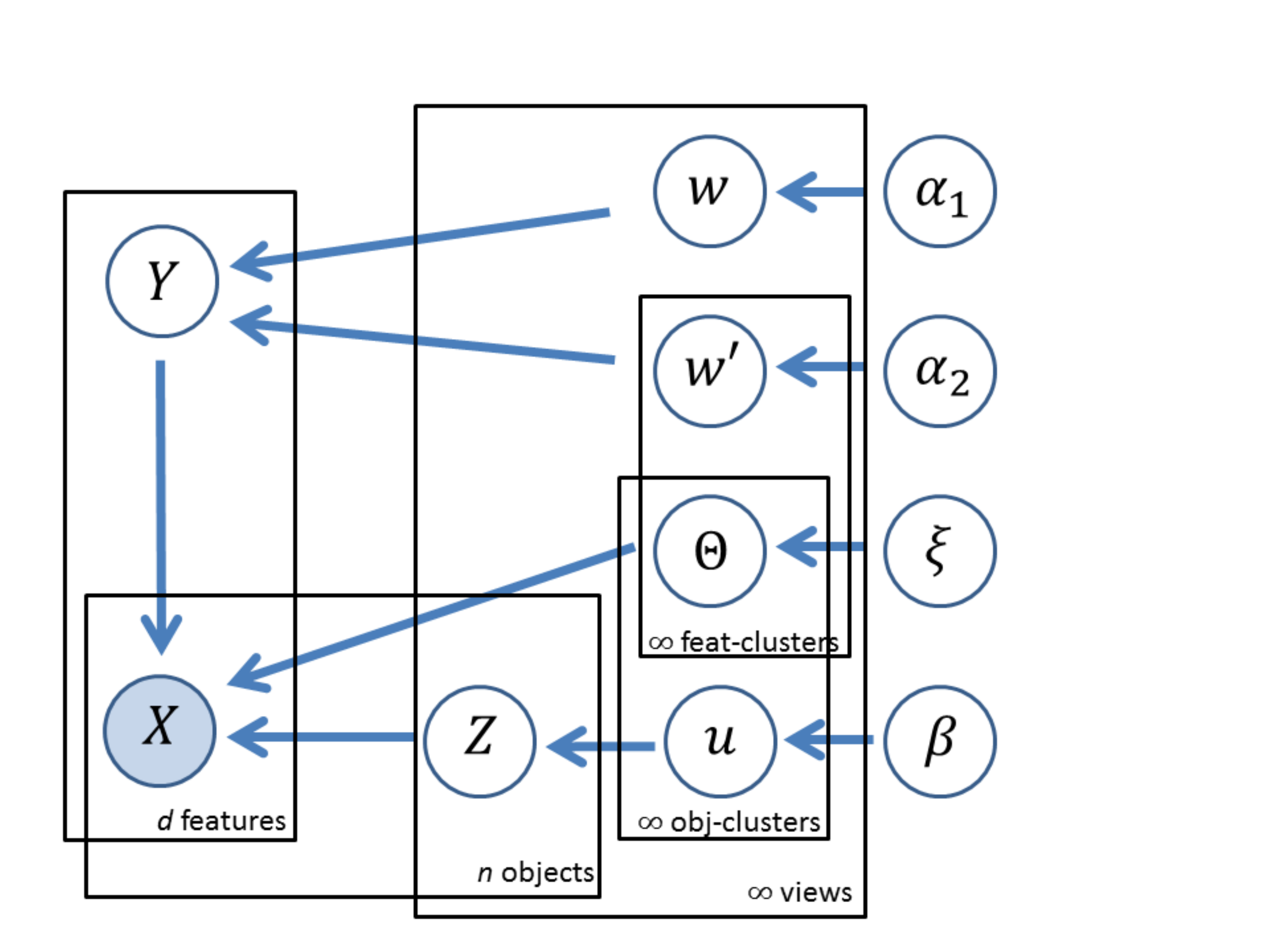} 
\vspace{-5mm}
\caption{\it Graphical model of relevant parameters in our multiple co-clustering model. Feat- and obj-cluster denotes feature and object cluster, respectively.  
Note that $\xi$ denotes all hyperparameters for distributions of parameters $\bb{\Theta}$. 
} 
\label{graphicalmodel}
\end{figure}

\vspace{5mm}
Our multiple clustering model is summarized in a graphical model of Figure~\ref{graphicalmodel}. It clarifies causal links among relevant parameters and a data matrix. 

\subsection{Likelihood and prior distribution}
We assume that each instance $X_{i, j}^{(m)}$ independently follows a certain distribution, 
conditional on $\bb{Y}$ and $\bb{Z}$. We denote $\bb{\theta}^{(m)}_{v, g, k}$
as parameters of distribution family $m$ in the cluster block of view $v$, feature cluster $g$ and object cluster $k$. Further denoting 
$\bb{\Theta}=\{\bb{\theta}^{(m)}_{v, g, k}\}_{v, g, k, m}$, the logarithm of complete-data likelihood of $\bb{X}$ is given by
\begin{eqnarray}
\no  &&\log p(\bb{X}|\bb{Y}, \bb{Z}, \bb{\Theta}) 
 = \sum_{m, v, g, k, j, i}\mathbb{I}(Y_{j, v, g}^{(m)}=1)\mathbb{I}(Z_{i, v, k}=1)
\log p(X_{i, j}^{(m)}|\theta_{v, g, k}^{(m)}), 
\label{like}
\end{eqnarray}
where $\mathbb{I}(x)$ is an indicator function, i.e, returning 1 if $x$ is true, and 0 otherwise. 
Note that the complete-data likelihood is not directly associated with $\bb{w}=\{w_v\}_v$, $\bb{w}'=\{ {w'}_{g, v}^{(m)}\}_{g, v}$ and $\bb{u}=\{u_{k,v}\}_{k, v}$.
The joint prior distribution for parameters is given by
\begin{eqnarray*} 
p(\bb{w})p(\bb{w}')p(\bb{Y}|\bb{w}, \bb{w}')p(\bb{u})p(\bb{Z}|\bb{u})p(\bb{\Theta}).
\end{eqnarray*}

\subsection{Variational Inference}
As in \cite{guan2010variational}, we use variational Bayes EM for MAP (maximum a posteriori) estimation of $\bb{Y}$
and $\bb{Z}$. The logarithm of the marginal likelihood $p(\bb{X})$ is approximated using Jensen's inequality \cite{jensen1906fonctions}:
\begin{eqnarray}
  \log p(\bb{X}) 
 &\geq & \int q(\bb{\phi})\log \frac{p(\bb{X}, \bb{\phi})}{q(\bb{\phi})} d \bb{\phi} = \mathcal{L}(q(\bb{\phi})),
 \label{variinfer}
\end{eqnarray}
where $q(\bb{\phi})$ is an arbitrary distribution for parameters $\bb{\phi}$. It can be shown that the difference between the left and right sides is given by the Kullback-Leibler divergence between $q(\bb{\phi})$ and $p(\bb{\phi}|\bb{X})$, i.e., $\mathbb{KL} (q(\bb{\phi}), p(\bb{\phi}|\bb{X}))$. Hence, our approach of choosing $q(\bb{\phi})$ is to minimize $\mathbb{KL} (q(\bb{\phi}), p(\bb{\phi}|\bb{X}))$, which is tractable to evaluate. In our model, we choose $q(\bb{\phi})$ that is factorized over different parameters (mean field approximation):
\begin{eqnarray*}
 q(\bb{\phi}) 
   &=& q_{\bb{w}}(\bb{w})q_{\bb{w}'}(\bb{w}')q_{\bb{Y}}(\bb{Y})q_{\bb{u}}(\bb{u})q_{\bb{Z}}(\bb{Z})
q_{\bb{\Theta}}(\bb{\Theta}),
\end{eqnarray*}
where each $q(\cdot)$ is further factorized over subsets of parameters, $w_{v}$, 
 ${w'}_{g, v}^{(m)}$, $\bb{Y}_{j \cdot \cdot }^{(m)}$, $u_{k, v}$, $\bb{Z}_{i, v \cdot}$ and 
$\bb{\theta}^{(m)}_{v, g, k}$.

In general, the distribution $q_i(\phi_i)$ that minimizes $\mathbb{KL}(\prod_{l=1}^{L}q_l(\phi_l), p(\bb{\phi}|\bb{X}))$ is given by
\begin{eqnarray*}
  q_i(\phi_i) \propto  \exp \{\mathbb{E}_{-q_{i}({\phi})}\log p(\bb{X}, \bb{\phi})\},
\end{eqnarray*}
where  $\mathbb{E}_{-q_{i}({\phi})}$ denotes averaging with respect to $\prod_{l \neq i}q_l(\phi_l)$ \cite{Mur2012}. 
Applying this property to our model, it can be shown that
\begin{eqnarray*}
q_{\bb{w}}(\bb{w}) &=& \prod _{v=1}^{V} \mbox{Beta}(w_v|\gamma _{v, 1}, \gamma_{v, 2})\\
q_{\bb{w}'}(\bb{w}')&=& \prod_{m=1}^{M} \prod_{v=1}^{V} \prod_{g=1}^{G}\mbox{Beta}(w_{g, v}^{(m)}|\gamma _{g, v, 1}^{(m)}, \gamma_{g,v, 2}^{(m)})\\
q_{\bb{Y}}(\bb{Y}) &=& \prod_{m=1}^{M} \prod_{j=1}^{d^{(m)}} \mbox{Mul}(\bb{Y}_{j \cdot \cdot}^{(m)}|
{\bb{\tau}}_{j}^{(m)}) \\
q_{\bb{u}}(\bb{u}) &=& \prod_{v=1}^V \prod_{k=1}^K \mbox{Beta}(u_{g, v}|\gamma_{k, v, 1}, \gamma_{k, v, 2})\\
q_{\bb{Z}}(\bb{Z})&=& \prod_{v=1}^V \prod_{i=1}^n \mbox{Mul}(\bb{Z}_{i, v \cdot}| \bb{\eta}_{i, v})\\
\log q_{\bb{\Theta}}(\bb{\Theta}) &=& \sum_{m, v, g, k, j, i} \tau_{j, v, g}^{(m)} \eta_{i, v, k}\log p(X_{i, j}^{(m)}|
\bb{\theta}^{(m)}_{v, g, k})
 +\sum_{m, v, g, k} \log p(\bb{\theta}^{(m)}_{v, g, k})
+ \mbox{constant},
\end{eqnarray*}
where the hyperparameters except for $q_{\bb{\Theta}}(\bb{\Theta})$ are given by
\begin{eqnarray}
\no \gamma_{v, 1} &=& 1 + \sum_{m=1}^M\sum_{g=1}^G \sum_{j=1}^{d^{(m)}} \tau^{(m)}_{j,g,v}\\
\no \gamma_{v, 2} &=& \alpha_1 + \sum_{m=1}^M \sum_{t=v+1}^{V}\sum_{g=1}^G\sum_{j=1}^{d^{(m)}} \tau_{j, g, t}^{(m)}\\
\no \gamma_{g, v, 1}^{(m)} &=& 1 + \sum_{j=1}^{d^{(m)}}\tau_{j, g, v}^{(m)}\\
\no \gamma_{g, v, 2}^{(m)} &=& \alpha_2 + \sum_{t=g+1}^{G} \sum_{j=1}^{d^{(m)}}\tau_{j, t, v}^{(m)}\\
\no \gamma_{k, v, 1} &=& 1 + \sum_{i=1}^n \eta_{i, v, k} \\
\no \gamma_{k, v, 2} &=& \beta + \sum_{t=k+1}^K  \sum_{i=1}^{n} \eta_{i, v, t} \\
\no \log \tau_{j, g, v}^{(m)} &= & \sum_{k=1}^K \sum_{i=1}^n 
 \eta_{i, v, k} \mathbb{E}_{q(\bb{\theta})}\Big [\log p(X_{i, j}^{(m)} | 
\bb{\theta}_{v, g, k}^{(m)})\Big ]\\
\no && + \psi(\gamma_{v, 1})-\psi(\gamma_{v, 1} + \gamma_{v, 2})\\
\no && + \sum_{t=1}^{v-1}\{\psi (\gamma_{t, 2})-\psi(\gamma_{t, 1} + \gamma_{t, 2})\}\\
\no  && +\psi (\gamma_{g, v, 1}^{(m)})- \psi (\gamma_{g, v, 1}^{(m)} + \gamma_{g, v, 2}^{(m)} )\\
\no && +  \sum_{t=1}^{G-1}\{\psi(\gamma_{t, v, 2}^{(m)})-\psi(\gamma_{t, v, 1}^{(m)} + \gamma_{t, v, 2}^{(m)})\} \\
\no && +  \mbox{constant} \\
\no \log \eta_{i, v, k} &= & \sum_{m=1}^M \sum_{g=1}^{G}\sum_{j=1}^{d^{(m)}}
\tau_{j, g, v}^{(m)} \mathbb{E}_{q(\bb{\theta})}\Big [\log p(X_{i, j}^{(m)} | 
\bb{\theta}_{v, g, k}^{(m)})\Big ] \\
\no && +\psi (\gamma_{k, v, 1})- \psi (\gamma_{k, v, 1} + \gamma_{k, v, 2} )\\
\no && + \sum_{t=1}^{K-1}\{\psi(\gamma_{t, v, 2})-\psi(\gamma_{t, v, 1} + \gamma_{t, v, 2})\} \\
 && +  \mbox{constant},
 \label{update}
\end{eqnarray}
where $\mathbb{E}_{q(\bb{\theta})}$ denotes averaging with respect to the corresponding 
$q(\bb{\theta})$ of $\bb{\theta}_{v, g, k}^{(m)}$; $\psi (\cdot)$ denotes the digamma function defined as the first derivative of logarithm of gamma function. 
Note that 
$\tau_{j, g, v}^{(m)}$ is normalized over pairs $(g, v)$ for each pair $(j, m)$, while
$\eta_{i, v, k}$ normalized over $k$ for each pair of $(i, v)$.   
Observation models and priors of parameters $\bb{\Theta}$ are specified in the following section.

\subsection{Observation models} \label{obsmodel}
For observation models, we consider Gaussian, Poisson, and Categorical/multinomial distributions. For each cluster block, we fit a univariate distribution of these families with the assumption that features within the cluster block are independent. We assume conjugate priors for the parameters of these distribution families. Variational inference and updating equations are basically the same as in \cite{guan2010variational} (See Appendix~\ref{appenobs}).

\subsection{Algorithm} \label{algorithm}
With the updating equations of the hyperparameters, the variational Bayes EM proceeds as follows. First, we randomly initialize $\{\bb{\tau}^{(m)}\}_m$ and $\{\bb{\eta}_v\}_v$, and then alternatively
update the hyperparameters until the lower bound 
 $\mathcal{L}(q(\bb{\phi}))$ in Eqs.(\ref{variinfer}) converges. This yields possible optimal distributions $q_{\bb{Y}}(\bb{Y})$ and $q_{\bb{Z}}(\bb{Z})$, which immediately gives MAP estimates for $\bb{Y}$ and $\bb{Z}$. We repeat this procedure a number of times, and choose the best solution with the largest lower bound. 
The algorithm is outlined in Algorithm~\ref{algo}.
Note that the lower bound  $\mathcal{L}(q(\bb{\phi}))$ is given by
\begin{eqnarray}
    \mathcal{L}(q(\bb{\phi}))= 
     \int q(\bb{\phi})\log p(X|\bb{\phi})d\bb{\phi} -\mathbb{KL}(q(\bb{\phi}), p(\phi)),
     \label{lowerbound}
\end{eqnarray}
where both terms on the right side can be derived in closed form. It can be shown that this monotonically increases as 
$q(\bb{\phi})$ is optimized.

\begin{algorithm}[tb]
   \caption{Variational Bayes EM for multiple co-clustering}
   \label{algo}
\begin{algorithmic}
   \STATE {\bfseries Input:} data matrices $\bb{X}^{(1)}, \ldots, \bb{X}^{(M)}$.
   \FOR{$s=1$ {\bfseries to} $S$}
   \STATE Randomly initialize $\{\bb{\tau}^{(m)}\}_m$ and $\{\bb{\eta}_v\}_v$.
   \REPEAT
   \STATE -Update the hyperparameters of relevant distribution families  for $q_{\bb{\Theta}}(\bb{\Theta})$.
   \STATE -Update the hyperparameters for $q_{\bb{w}}(\bb{w})$, $q_{\bb{w}'}(\bb{w}')$, 
   $q_{\bb{Y}}(\bb{Y})$, $q_{\bb{u}}(\bb{u})$, and $q_{\bb{Z}}(\bb{Z})$.
   \UNTIL{$L$ in Eqs.(\ref{lowerbound}) converges.}
   \STATE Keep $L(s)=L$
   \ENDFOR 
   \STATE $s^*=\mbox{argmax}_s L(s)$
   \STATE {\bfseries Output:} MAP for $\bb{Y}$ and $\bb{Z}$ in the run $s^*$.
\end{algorithmic}
\end{algorithm}
 
\subsection{Time complexity} \label{timecomplex}
For simplicity, we consider time complexity of our algorithm for a single run. If we assume that the number of required iterations for convergence is the same, the time complexity of the algorithm is equivalent to the number of operations for updating the relevant parameters. In that case, as can be seen in the updating equations in Eqs.(\ref{update}) and Appendix~\ref{appenobs}, the time complexity is just $O(nd)$ where $n$ and $d$ are the number of objects and the number of features (we fix the number of views and clusters). This enhances efficiency in applying our multiple co-clustering method to high-dimensional data. We return to this point in Section~\ref{complexity}  to compare other multiple clustering methods.

\subsection{Model representation}
Our multiple co-clustering model is sufficiently flexible to represent different clustering models because the number of views and the number of feature-/object clusters are derived in a data-driven approach. For instance, when the number of views is one, the model coincides with a co-clustering model; when the number of feature clusters is one for all views, it matches the restricted multiple clustering model. Furthermore, when the number of views is one and the number of feature clusters is the same as the number of features, it matches conventional mixture models with independent features. Moreover, our model can detect non-informative features that do not discriminate between object clusters. In such a case, the model yields a view in which the number of object clusters is one. The advantage of our model is to automatically detect such underlying data structures.

\subsection{Missing values} \label{missingvalue}
Our multiple co-clustering model can easily handle missing values. Suppose that the missing entries occur at random, which may depend on the observed data, but not the missing ones (i.e., MAR, missing at random). We can deal with such missing values in a conventional Bayesian way, in which missing entries are considered as stochastic parameters \cite{Gel2004}. In our model, this procedure is simply reduced to ignoring these missing entries when we update the hyperparameters. This is because (univariate) instances within a cluster block are assumed to be independent; hence the log-likelihood in Eqs.(\ref{like}) is given by
\begin{eqnarray*}
  &&\log p(\bb{X}^{obs}|\bb{Y}, \bb{Z}, \bb{\Theta})
   = \sum_{m, v, g, k, j, i} 
\mathbb{I}(Y_{j, v, g}^{(m)}=1)\mathbb{I}(Z_{i, v, k}=1)
\mathbb{I}((i, j)^{(m)} \in o)
\log p(X_{i, j}^{(m)}|\theta_{v, g, k}^{(m)}),
\end{eqnarray*}
where $\mathbb{I}((i, j)^{(m)} \in o)$ is an indicator for the status of availability of the data cell 
of object $i$ and feature $j$ for distribution family $m$ (1 when it is available, and 0 otherwise); $\bb{X}^{obs}$ a subset of $\bb{X}$ that consists of the observed data only.

\section{Simulation study on synthetic data}\label{sectionsim}
\begin{table}[!h]
\caption{Summary of results of  simulation study on synthetic data: Recovery of true object cluster structure and views evaluated in terms of mean values of adjusted Rand Index.}
\begin{center}
\small
\begin{threeparttable}
\begin{tabular}{lc ||ccc|ccc}
\multicolumn{2}{c}{}  & \multicolumn{3}{c}{Object clustering}  &  \multicolumn{3}{c}{Views}\\
\hline
 \multicolumn{1}{c}{Factors} & & \textbf{Mul} & Co & rMul & \textbf{Mul} & Co & rMul \\
\hline
Number     & 20 & \textbf{0.30}   & 0.21  & 0.13  & 0.05 & \textbf{0.27}  & 0.11 \\
 of objects            & & (0.01) & (0.00) &  (0.01) & (0.01) & (0.00) & (0.00) \\
                & 50 & \textbf{0.81} & 0.23 & 0.51 & \textbf{0.75} & 0.48 & 0.37 \\
                 && (0.01) & (0.00) & (0.01) & (0.01) & (0.00) & (0.01)\\ 
                 & 100 & \textbf{0.84} & 0.21 & 0.69 & \textbf{0.77} & 0.54 & 0.43 \\
                && (0.01) & (0.00) & (0.01) & (0.01) & (0.00) & (0.01) \\
\hline
 Number of    & 10 & \textbf{0.46} & 0.21 & 0.26 & 0.29  & \textbf{0.33} & 0.22 \\
 features                   && (0.01) & (0.00) & (0.01) & (0.01) & (0.01) & (0.01) \\
 & 50 & \textbf{0.75} & 0.22 & 0.49 & \textbf{0.66} & 0.47 & 0.33\\
                  && (0.01) & (0.00) & (0.01) & (0.01) & (0.00) & (0.01)\\
        & 100 & \textbf{0.73} & 0.21 & 0.58 & \textbf{0.62} & 0.49 & 0.37 \\
        && (0.01) & (0.00) & (0.01) & (0.01) & (0.00) & (0.01)\\
\hline
 Ratio of  & 0 & \textbf{0.70} & 0.22 & 0.49 & \textbf{0.58} & 0.45 & 0.33 \\
 missings  && (0.01) & (0.00) & (0.01) & (0.01) & (0.00) & (0.01) \\
& 0.1 & \textbf{0.65} & 0.22 & 0.44 & \textbf{0.53} & 0.43 & 0.30\\
&& (0.01) & (0.00) & (0.01) & (0.01) & (0.01) & (0.01) \\
 & 0.2 & \textbf{0.59} & 0.21 & 0.40 & \textbf{0.48} & 0.41 & 0.28\\
 && (0.01) & (0.00) & (0.01) & (0.01) & (0.01) & (0.01) \\
\hline
\end{tabular}
\begin{tablenotes}
\small \it
\item[](a) Mul, Co, and rMul denote our multiple co-clustering, co-clustering and restricted multiple clustering methods.
\item[](b) Digits without parenthesis denotes mean values of adjusted Rand Index over 9$\times$100 = 900 datasets for a corresponding factor. 
\item[] (c) Digits in parenthesis denote standard errors of the mean.
\item[] (d) The digits for the best performance among the three methods are in bold. 
\end{tablenotes}
\end{threeparttable}
\end{center}
\label{tableres}
\end{table}

\begin{figure}[!h]
\centering
\includegraphics[scale=0.4]{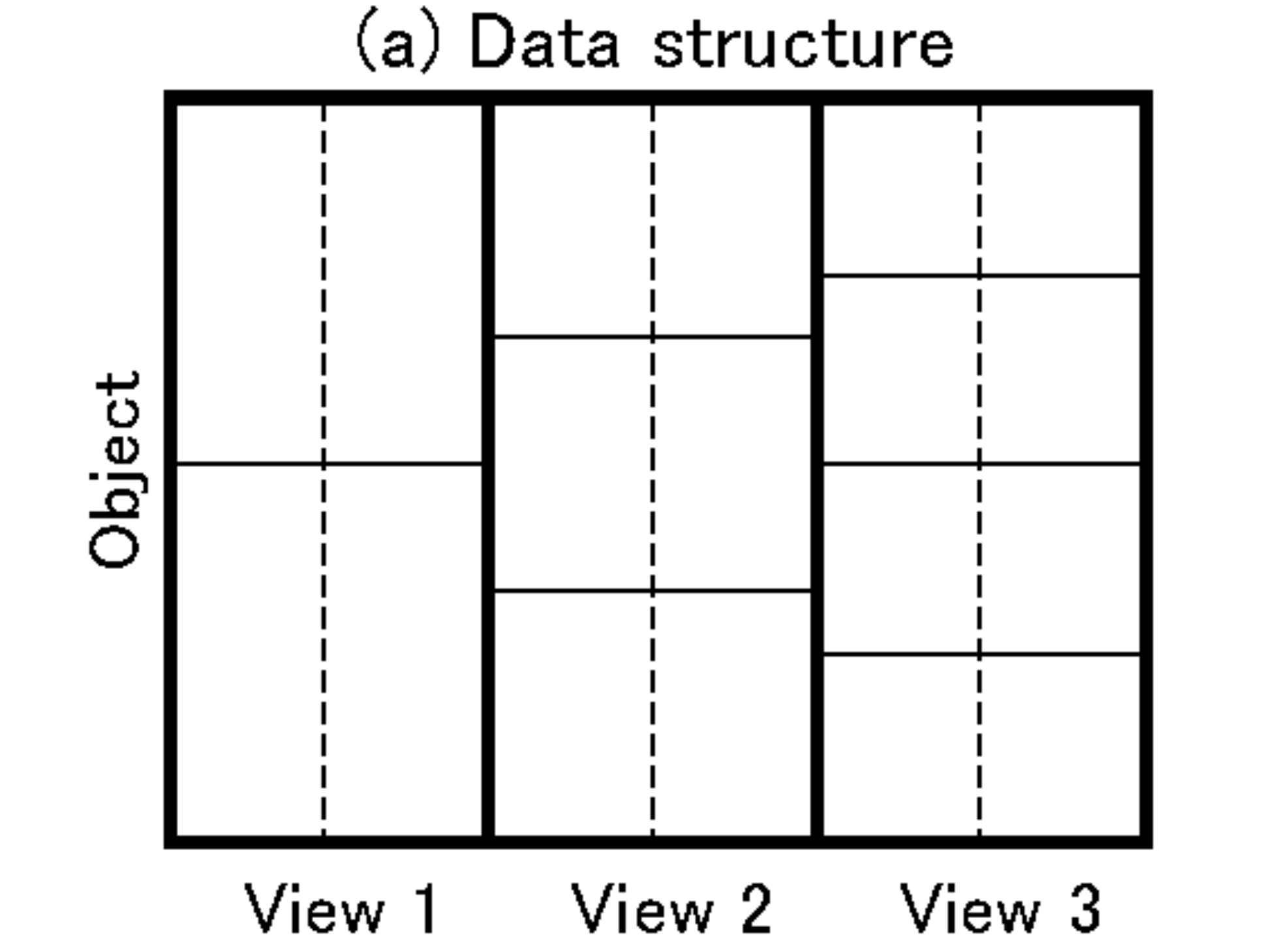}%
\includegraphics[scale=0.4]{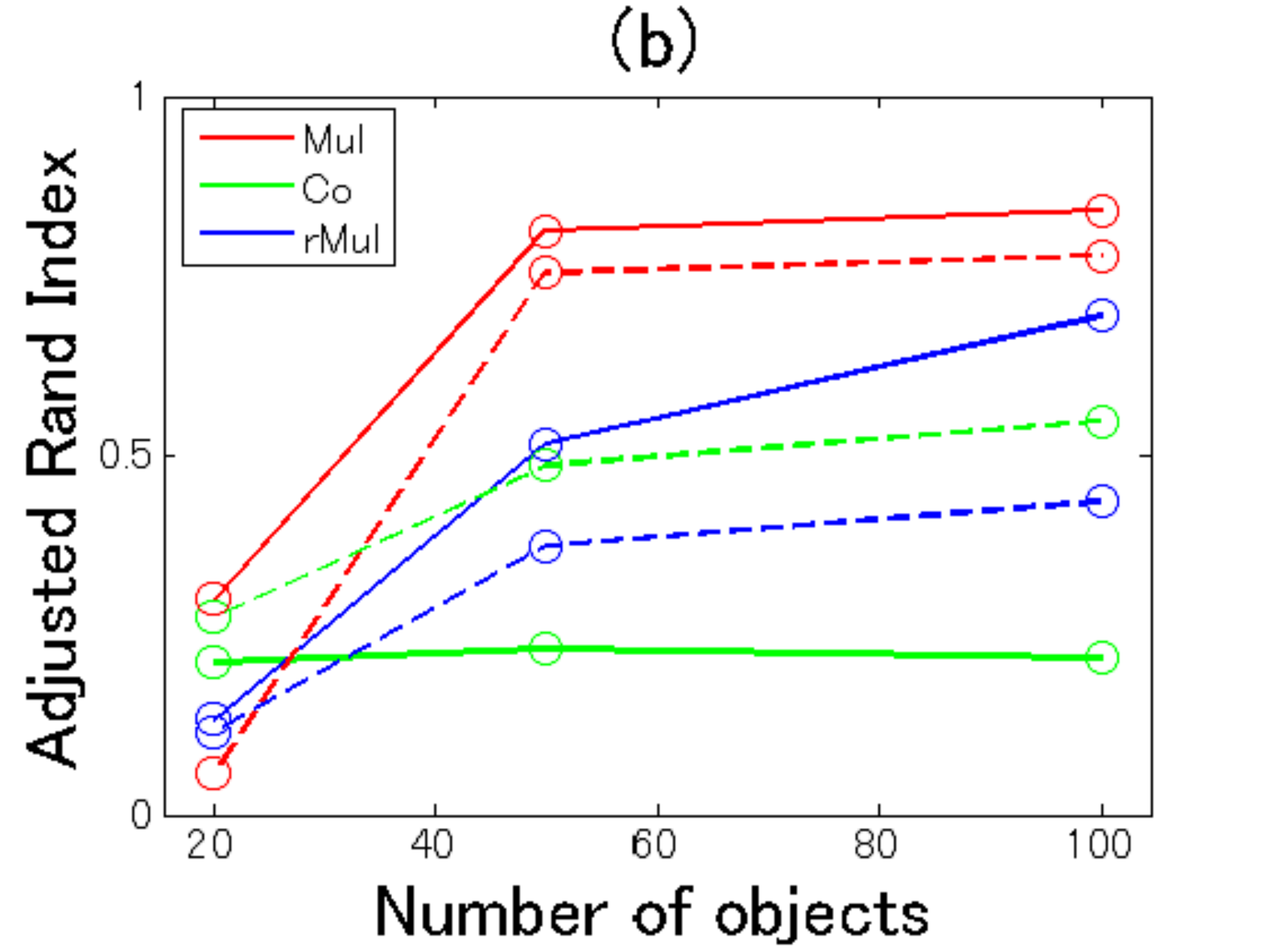} %
\includegraphics[scale=0.4]{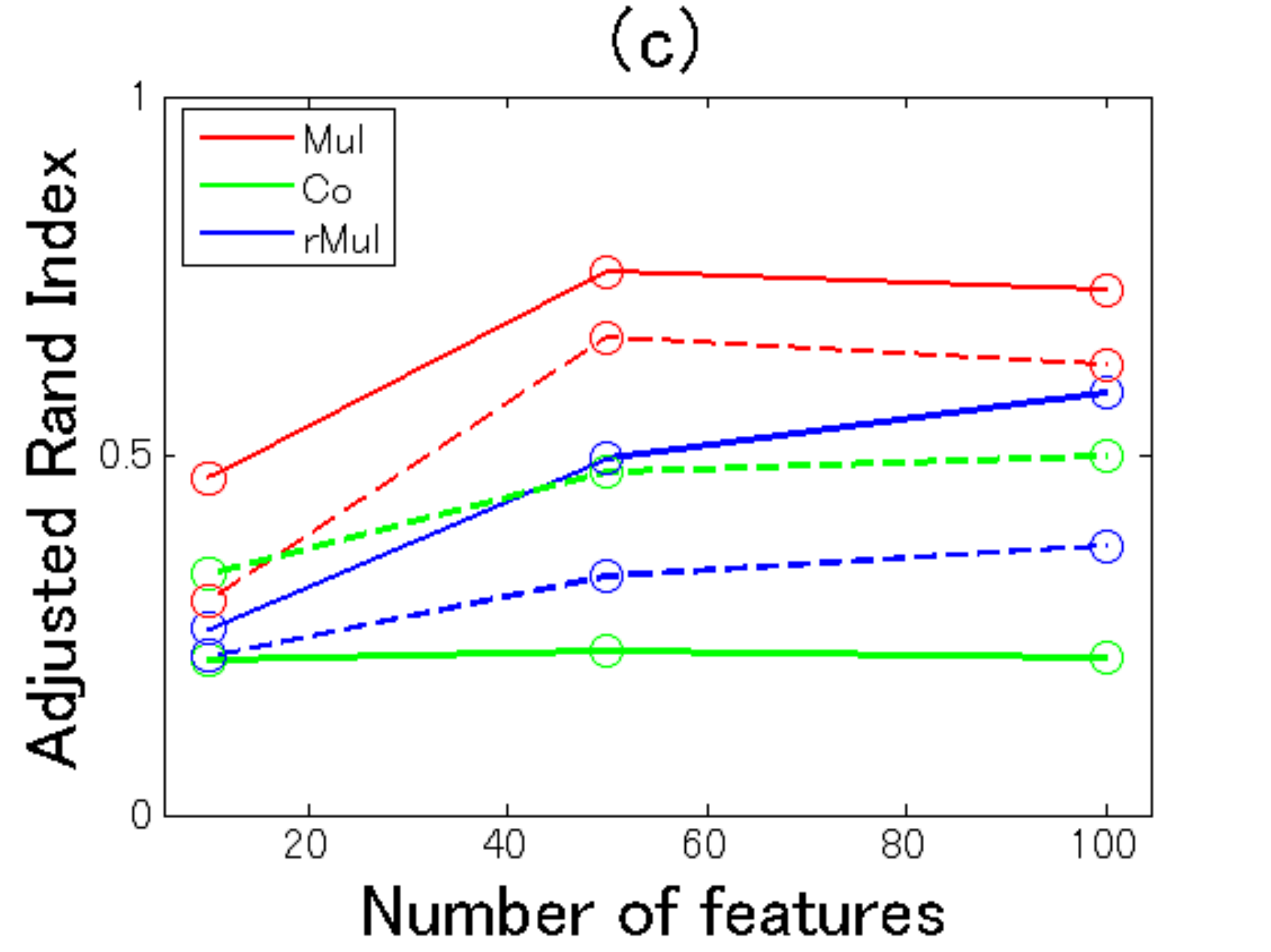} 
\includegraphics[scale=0.4]{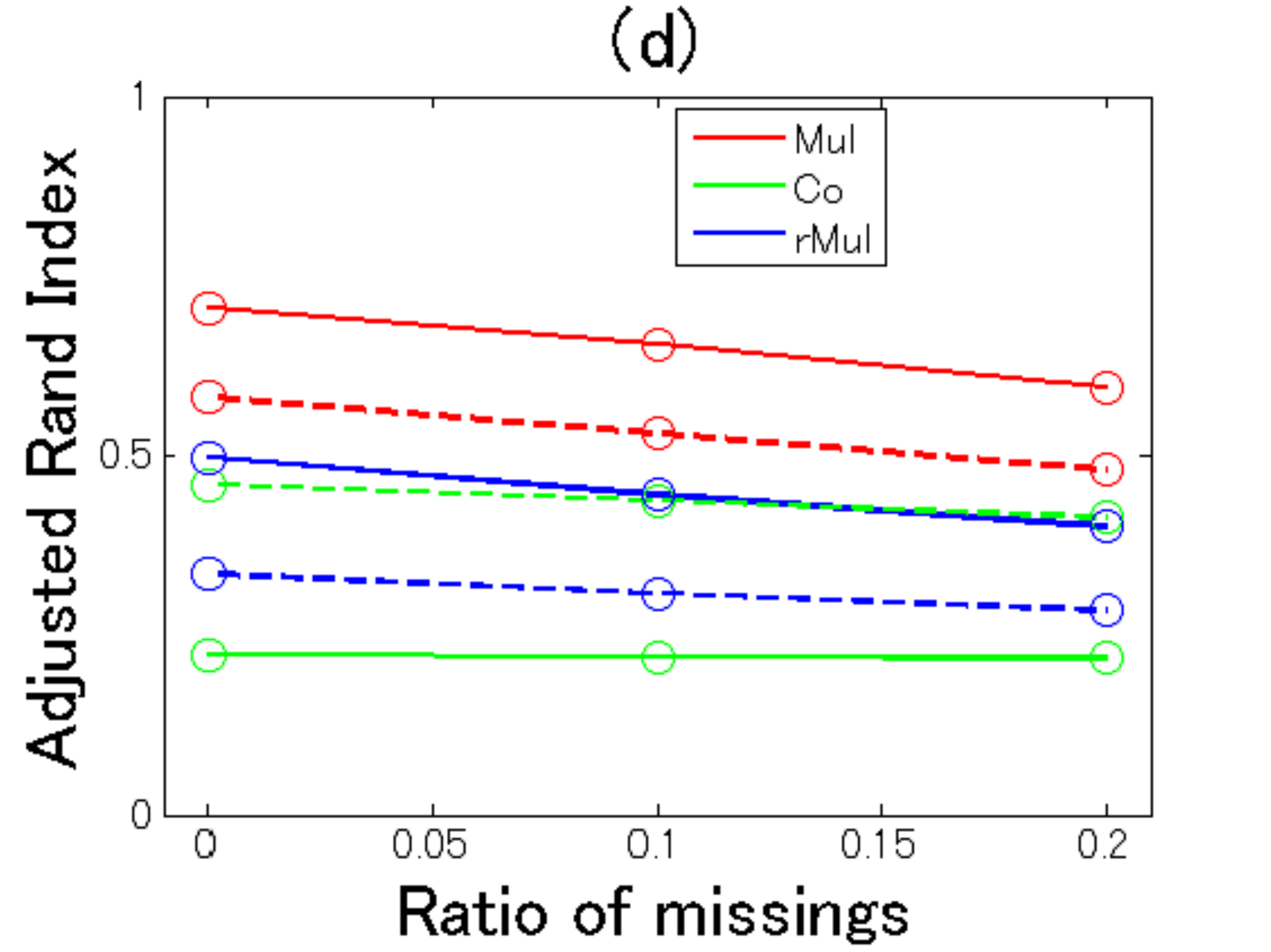} 
\caption{\it \small
Panel (a): Data structure for the simulation study. Each view has two feature clusters (separated by a dashed line) for each type of features of Gaussian, Poisson and Categorical. For Gaussian, 
means are set to (0, 4 ; 1, 3) ((0, 4) for top left and right cluster blocks, (1, 3) for bottom left and right cluster blocks) for view 1; (0, 5 ; 1, 4 ; 2, 3) for view 2; (0, 6 ; 1, 5 ; 2, 4 ; 3, 3) for view 3. The standard deviation is fixed to one. Similarly, for Poisson, the parameter $\lambda$ is set to (1, 2 ; 2, 1), (1, 3 ; 2, 2 ; 3, 1), (1, 4 ; 2, 3 ; 3, 2 ; 4,1). For categorical (binary), probability for success is (0.1, 0.9 ; 0.1, 0.9),(0.1, 0.9 ; 0.5, 0.5; 0.9, 0.1), (0.1, 0.9 ; 0.4, 0.6 ; 0.6, 0.4 ; 0.9, 0.1). Panels (b)-(d): Performance of the multiple co-clustering method ('Mul', red), the co-clustering method ('Co', green), and the restricted multiple clustering method ('rMul', blue).
Solid lines are for recovery of object cluster solutions, while dashed lines are for recovery of views.  
The results are summarized with respect to the number of objects (b), the number of features (c) and the proportions of missing entries (d).
}  
\label{simres}
\end{figure}

In this section, we report on a simulation study to evaluate the performance of our method. To the best of our knowledge, there is no algorithm in the literature that allows mixing of different types of features, as we have so far modeled. Hence, we compare the performance of our multiple co-clustering method only with co-clustering and restricted multiple clustering methods, which we model to accommodate different types of features. We set the hyperparameters $\alpha_1$, $\alpha_2$, and $\beta$ relevant for generating views, feature clusters, and sample clusters in Section~\ref{hyperpara}  to one, and the hyperparameters relevant for observations models to those specified in Appendix~\ref{appenobs}. Note that we use this setting for further application of our method in Section~
\ref{section application}. 
For data generation, we fixed the number of views to three and the number of object clusters to two, three, and four in views 1-3, respectively. The number of feature clusters was set to two in all views (Figure~\ref{simres}a). We manipulated the number of features (per view \textit{and}  distribution family) (10, 50, 100), the number of objects (20, 50, 100), and the proportion of (uniformly randomly generated) missing entries (0, 0.1, 0.2). We included three types of mixtures of distributions: Gaussian, Poisson, and Categorical. Memberships of views were evenly assigned to features for each distribution family, and the feature and object cluster memberships were uniformly randomly allocated. The distribution parameters for each cluster block were fixed as in the legend of Figure~\ref{simres}. We generated 100 datasets for each setting, which resulted in $100\times 27$=2700  datasets.

We evaluated the performance of recovering the true cluster structure by means of an adjusted Rand index (ARI) 
\cite{hubert1985comparing}: When ARI is one, recovery of the true cluster structure is perfect. When ARI is close to zero, recovery is almost random. Specifically, we focused on recovery of memberships of views, and memberships of object clusters. Since the numbering of views is arbitrary, it is not straightforward to evaluate recovery of the true object cluster solutions (the correspondence between the yielded and the true object cluster solutions is not clear). Hence, to evaluate the performance of object cluster solutions, we first evaluated ARIs for all combinations of the true object clusters and yielded object cluster solutions, and then found the maximum ARI for each true object cluster solution. Lastly, we averaged the ARIs over views. In this manner, we evaluated the performance for the multiple co-clustering method and the restricted multiple clustering. The co-clustering method yields only a single object cluster solution; hence we averaged ARIs between the true object cluster solutions and this solution.

The performance of the multiple co-clustering method is reasonably good: performance of the recovery of views (red dashed line) and object clusters (red solid line) solutions improves as the number of objects increases 
(Table~\ref{tableres} and Figure~\ref{simres}b). Regarding the number of features, the performance improves as the number of features increases from 20 to 50, but there is no improvement from 50 to 100. This is possibly because in our simulation setting, each feature does not clearly discriminate between object clusters; hence, adding more features does not necessarily improve the recovery of views (hence, the recovery of object cluster solutions). Lastly, when the ratio of missing entries increases, the performance just becomes slightly worse, which suggests that our method is relatively robust to missing entries.

As a whole, the multiple co-clustering method outperforms the co-clustering and the restricted multiple clustering methods. The performance of the co-clustering method is poor because it does not fit the multiple clustering structure. On the other hand, the restricted multiple clustering method can potentially fit each object cluster structure; hence, it performs somewhat well in this regard (but, not for recovery of the true memberships of views).

\section{Application to real data} \label{section application}
To test our multiple co-clustering method on real data, we consider three datasets: facial image data, cardiac arrhythmia data, and depression data. For facial image and cardiac arrhythmia data, the (possible) true sample clustering label is available, which enables us to evaluate clustering performance of our multiple co-clustering method. We compare the performance with the restricted multiple clustering method and two state-of-the-art multiple clustering methods: the constrained orthogonal average link algorithm (COALA,  \cite{bae2006coala}) and the decorrelated $K$-means algorithm \cite{jain2008simultaneous}. These state-of-the-art methods aim to detect dissimilar multiple sample clustering solutions without partitioning of features. COALA is based on a hierarchical clustering algorithm, while decorrelated $K$-means is based on a $K$-means algorithm. The two methods also differ in the way to detect views: COALA iteratively identifies views while decorrelated $K$-means simultaneously does so. For both methods, we need to set the number of views and the number of sample clusters. In this experiment, we set these to the (possible) true numbers. For the depression data, no information is available on true cluster structure. Hence, we focus mainly on implications drawn from the data by our multiple co-clustering method, rather than evaluating the performance of recovery of true cluster structure.

\subsection{Facial image data} \label{face}
\begin{figure}[t]
\centering
\includegraphics[scale=0.3, trim=0mm 0mm 0mm 0mm]{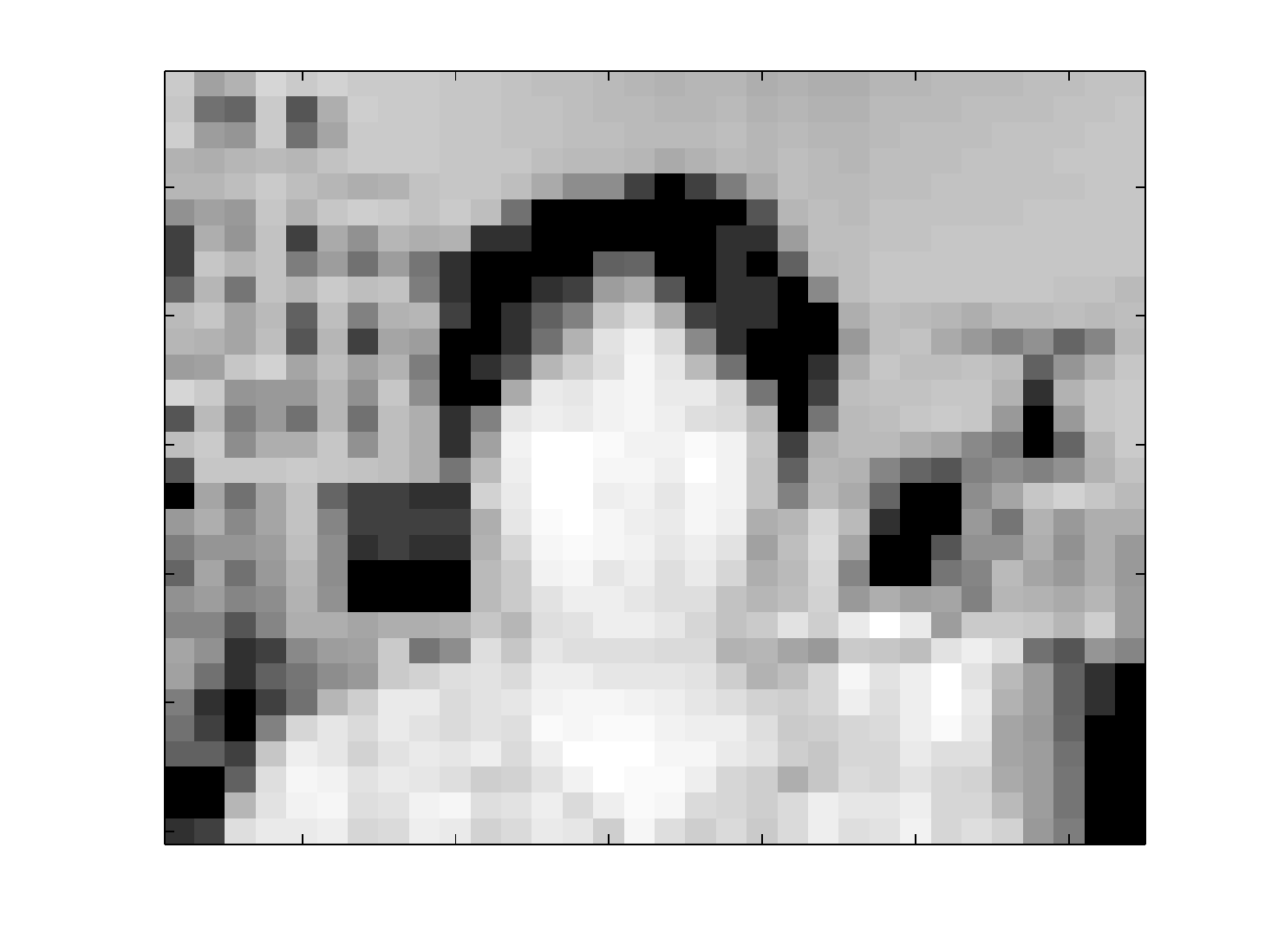} 
\includegraphics[scale=0.3, trim=0mm 0mm 0mm 0mm]{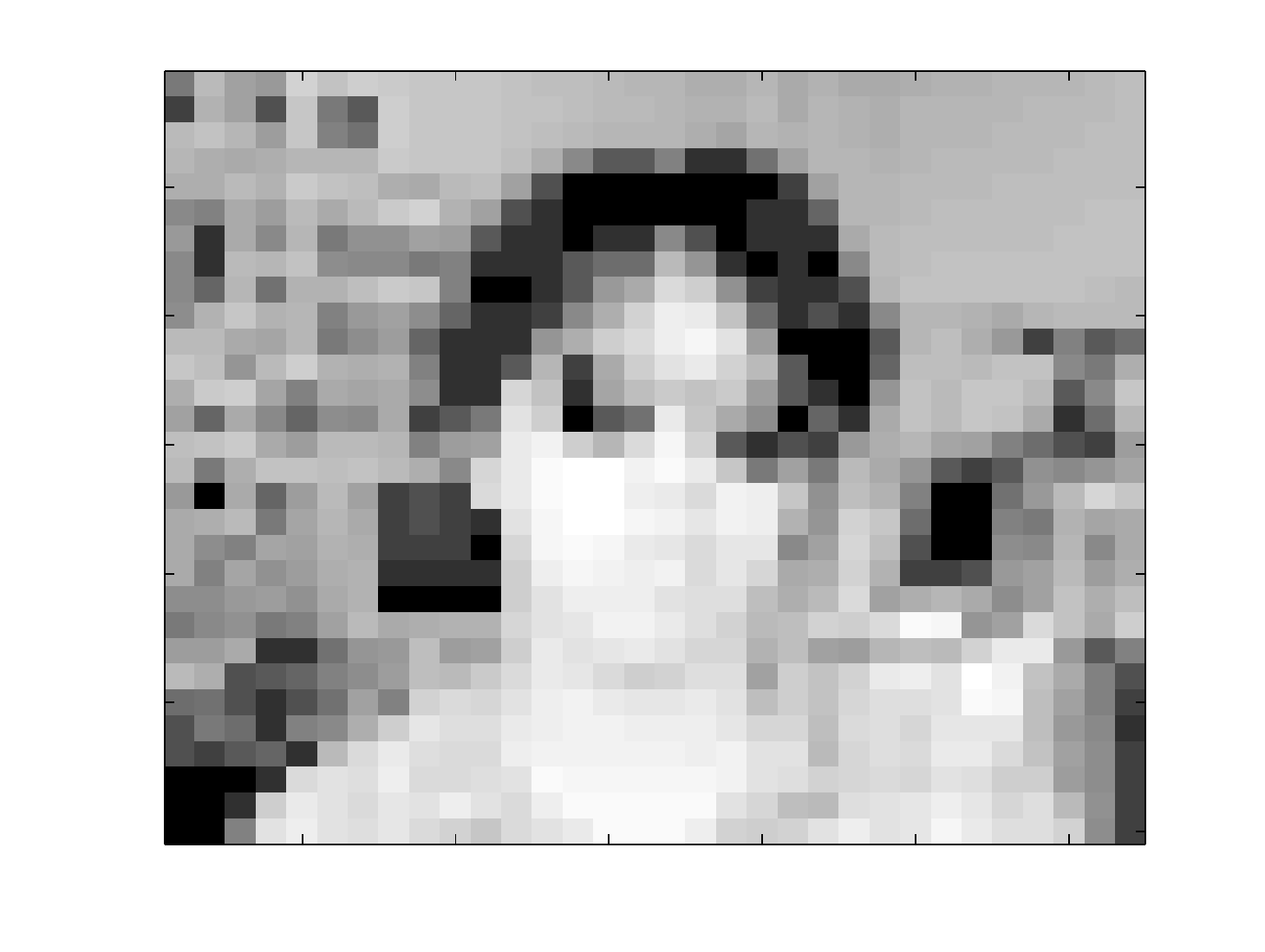} 
\includegraphics[scale=0.3, trim=0mm 0mm 0mm 0mm]{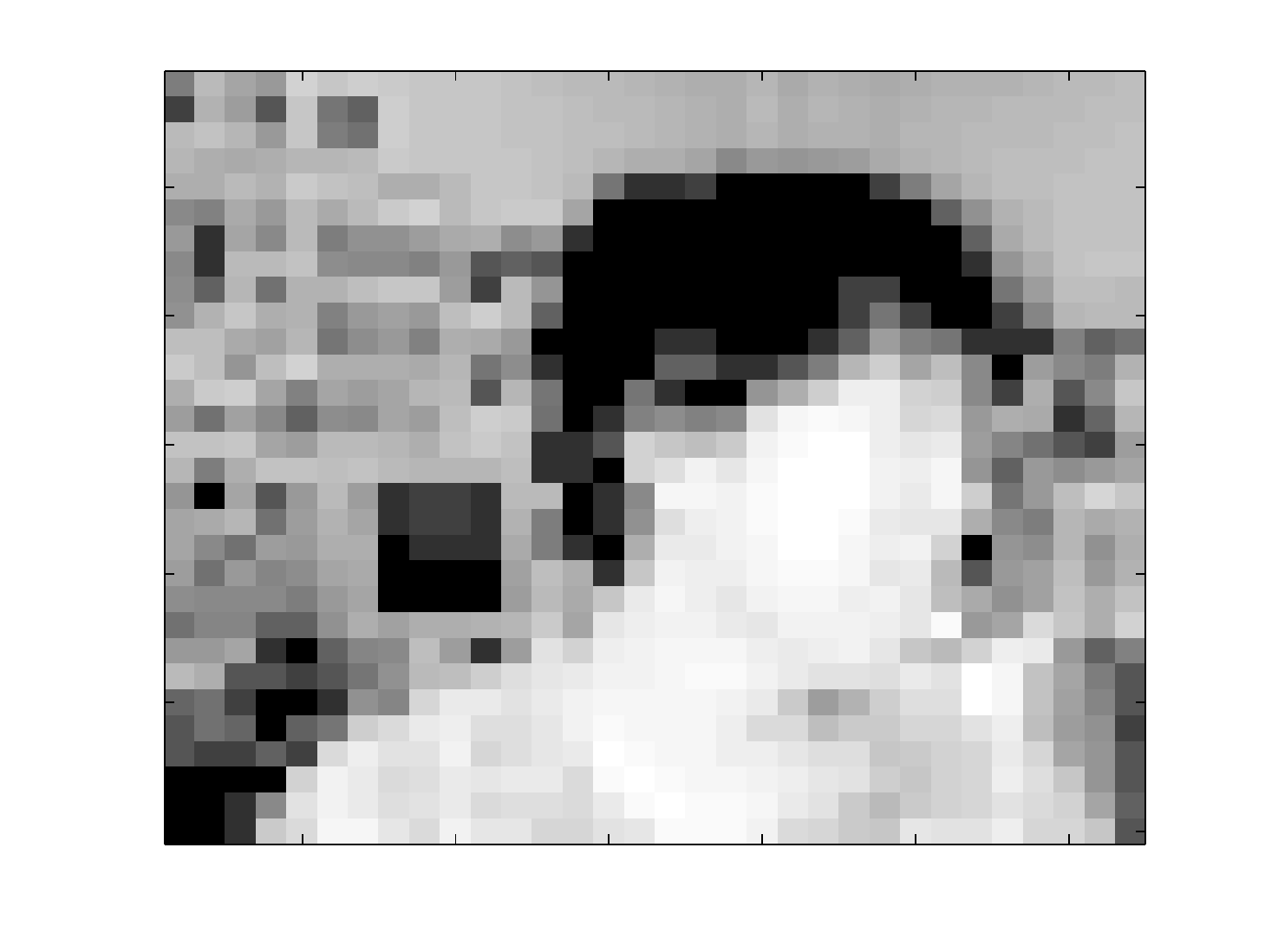} %
\includegraphics[scale=0.3, trim=0mm 0mm 0mm 0mm]{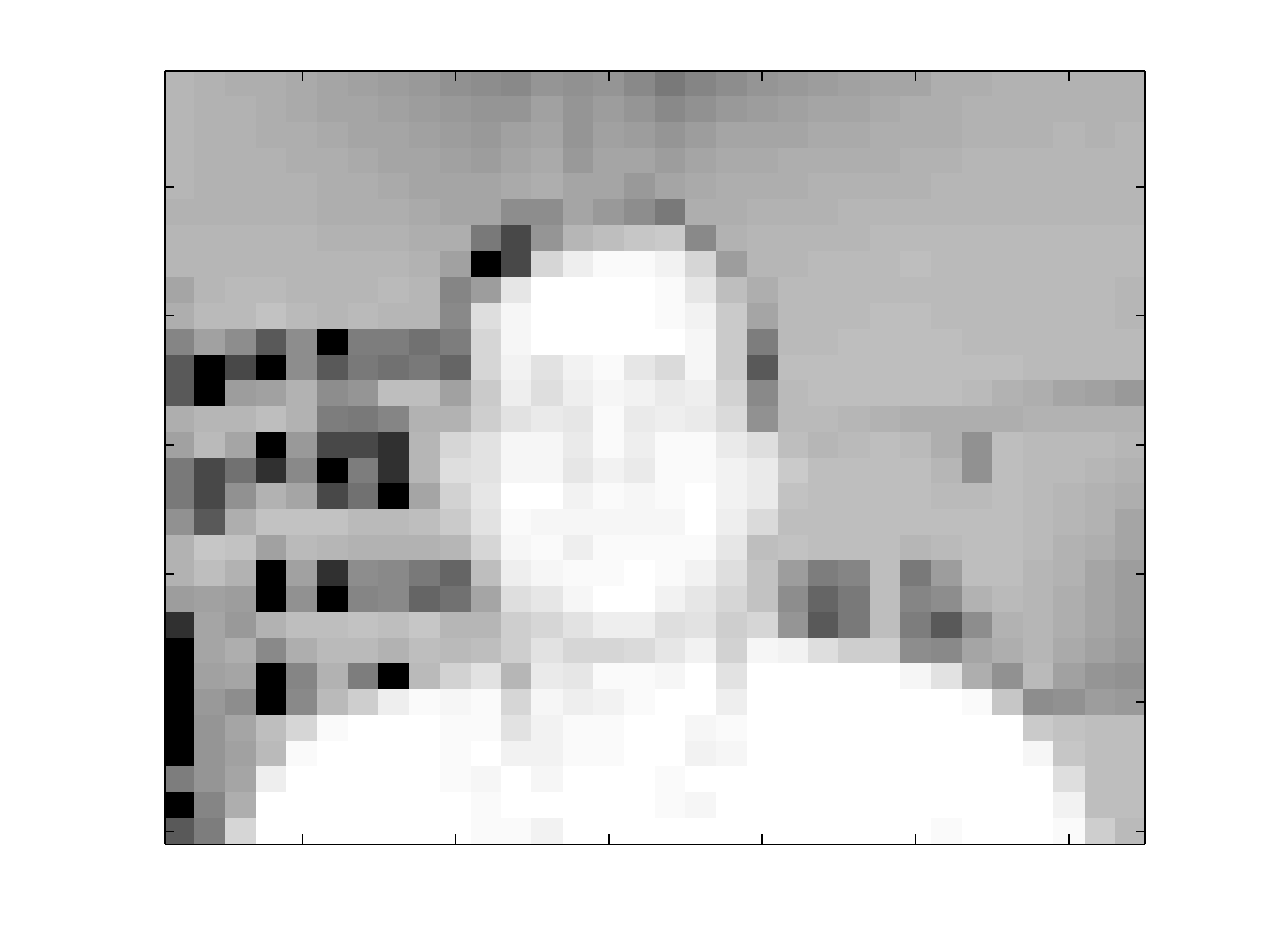} 
\includegraphics[scale=0.3, trim=0mm 0mm 0mm 0mm]{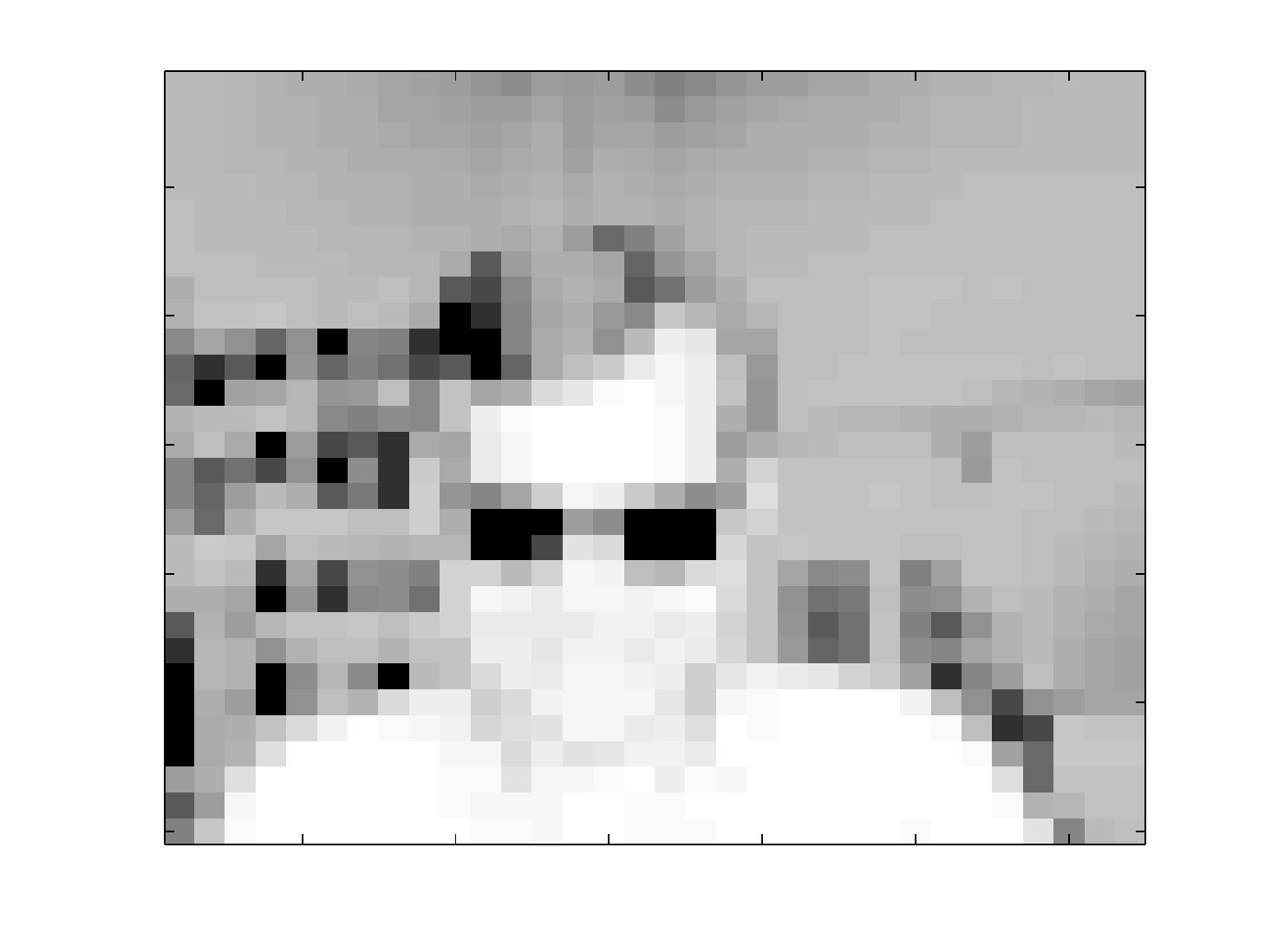} 
\includegraphics[scale=0.3, trim=0mm 0mm 0mm 0mm]{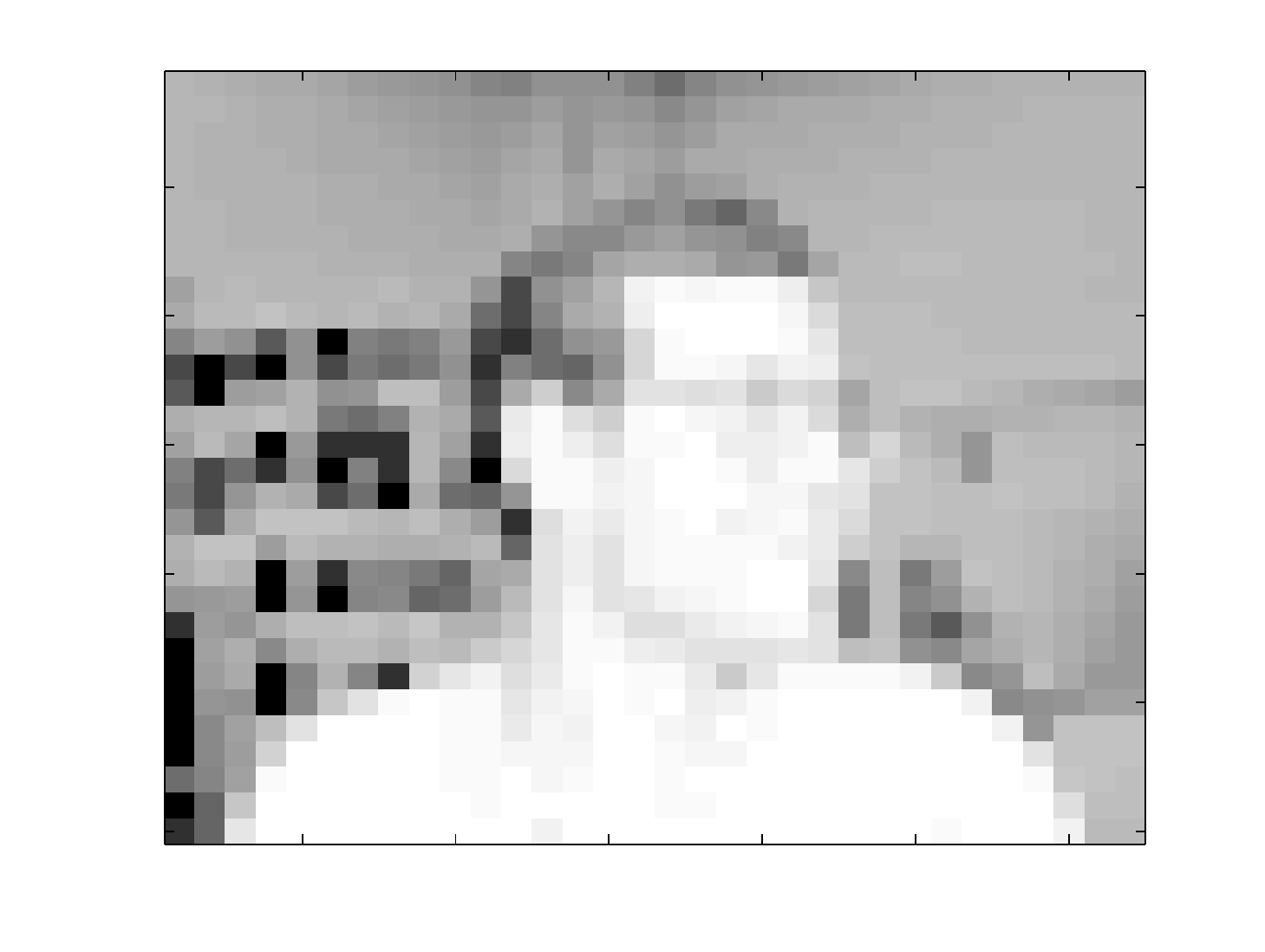} 
\caption{\it Samples from the facial image data: The first row represents person `an2i' with 
configurations of (no sunglasses, straight pose and neutral expression), (sunglasses, straight pose, angry expression) and (no sunglasses, left pose, happy expression) from left to right columns. The second row for person `at33' with the same patterns of configuration. 
}  
\label{image}
\end{figure}

The first dataset contains facial image data from the UCI KDD repository (\url{http://archive.ics.uci.edu/ml/datasets.html}), which consists of black and white images of 20 different persons with varying configurations (Figure \ref{image}): \textit{eyes} (wearing sunglasses or not), \textit{pose} (straight, left, right, up), and \textit{expression} (neutral, happy, sad, angry). This dataset served as a benchmark for multiple clustering in several papers \cite{guan2010variational, 
bae2006coala}. Here, we focus on the quarter-resolution images ($32 \times 30$) of this dataset, which results in 960 features. For simplicity, we consider two subsets of these images: data 1 consisting of a single person (named `an2i') with varying \textit{eyes}, \textit{pose} and \textit{expression} (data size: $32 \times 960$); data 2 consisting of two persons (in addition to `an2i', we include person `at33', data size: $64 \times 960$ ). We use these datasets without pre-processing.

The facial image data has multiple clustering structures that can be characterized by all of the features (global) or some of the features (local). Identification of persons (hereafter, \textit{useid}) may be related to global information of the image (all features), while \textit{eyes}, \textit{pose} and \textit{expression} are based on local information (a subset of features). Here, we focus only on \textit{pose}, which is a relatively easy aspect to detect \cite{niu2012nonparametric}. Since COALA and decorrelated $K$-means methods do not explicitly model relevant features for sample clustering, they can potentially capture a multiple clustering structure based on either global or local information of such a dataset. On the other hand, our multiple co-clustering model is based on a partition of features, which implicitly assumes that a possible sample clustering structure is based on non-overlapping local information. Our interest in this application is to examine the performance of our multiple clustering method using such data.

To evaluate performance, we focus only on sample clustering solutions. We base our evaluation criterion on recovery of structures of  \textit{useid} and \textit{pose} (\textit{useid} is applicable only for Data 2), which is measured by the maximum value of an adjusted Rand Index between the true sample structure in question and resulting sample clustering solutions. We discuss the results for each data in the following sub-sections.

\subsubsection{Results: Data 1}
\begin{figure}[t]
\centering
\includegraphics[scale=0.27, trim=0mm 0mm 0mm 0mm]{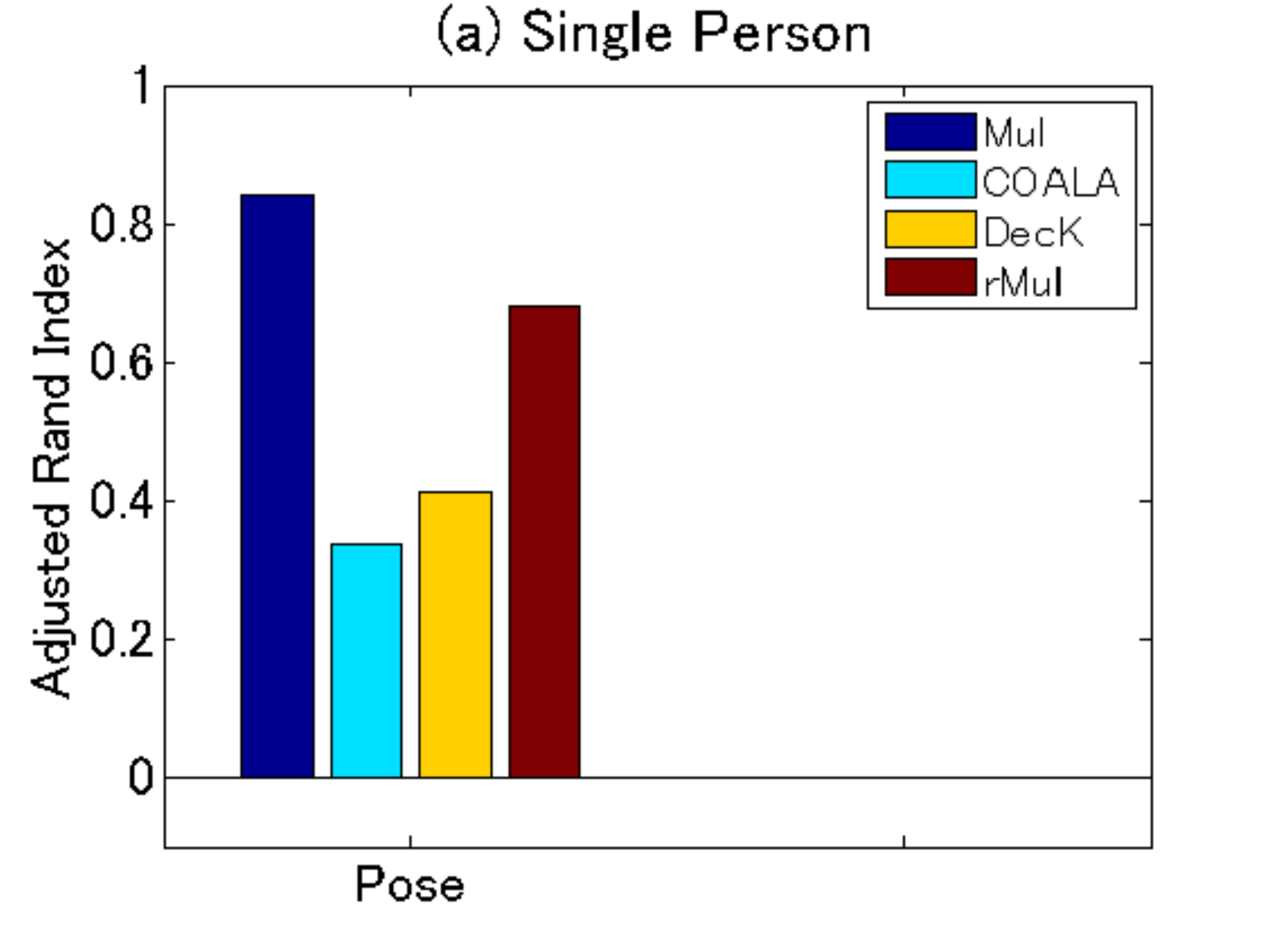} 
\includegraphics[scale=0.27, trim=0mm 0mm 0mm 0mm]{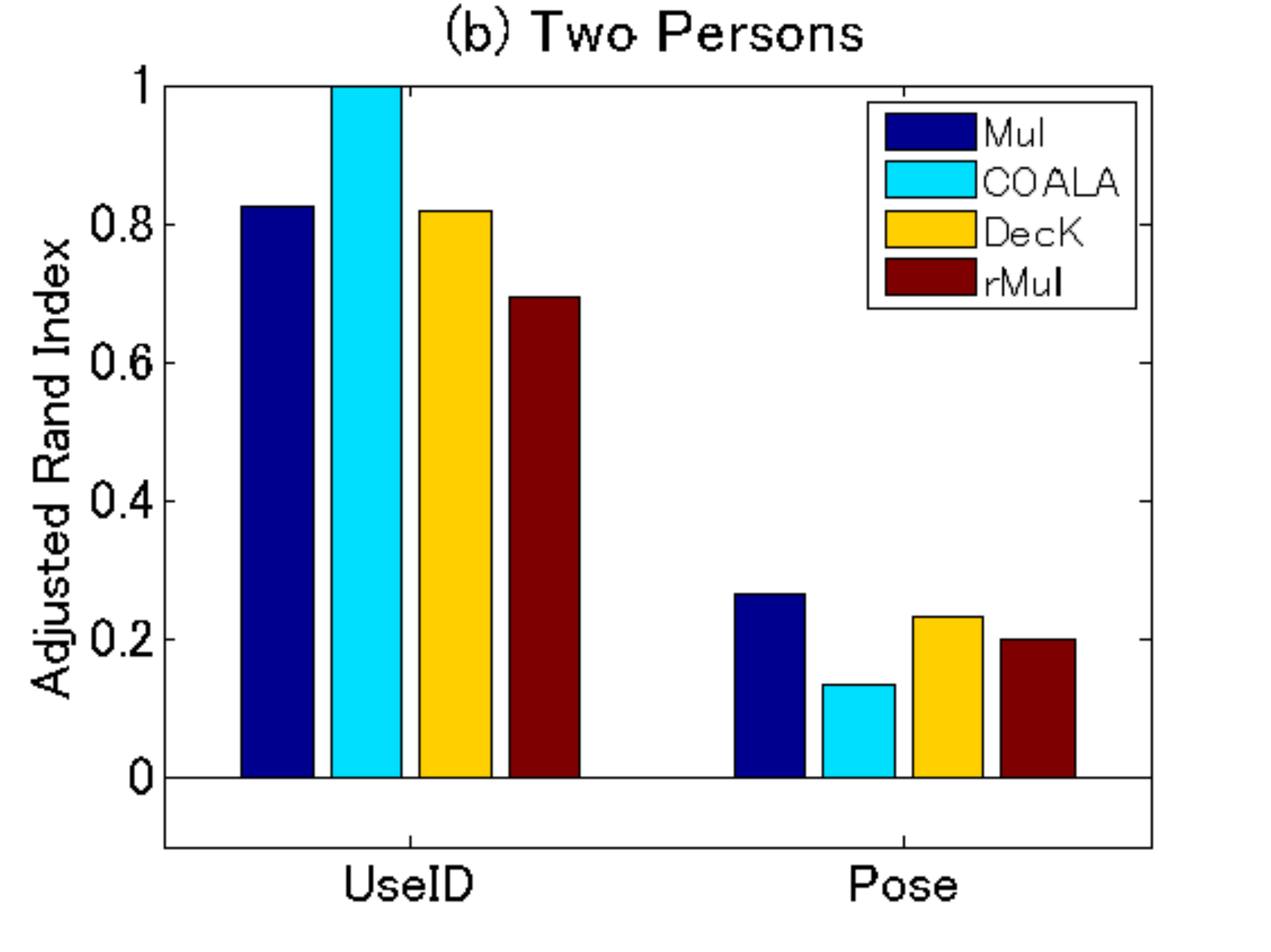} 
\includegraphics[scale=0.27, trim=0mm 0mm 0mm 0mm]{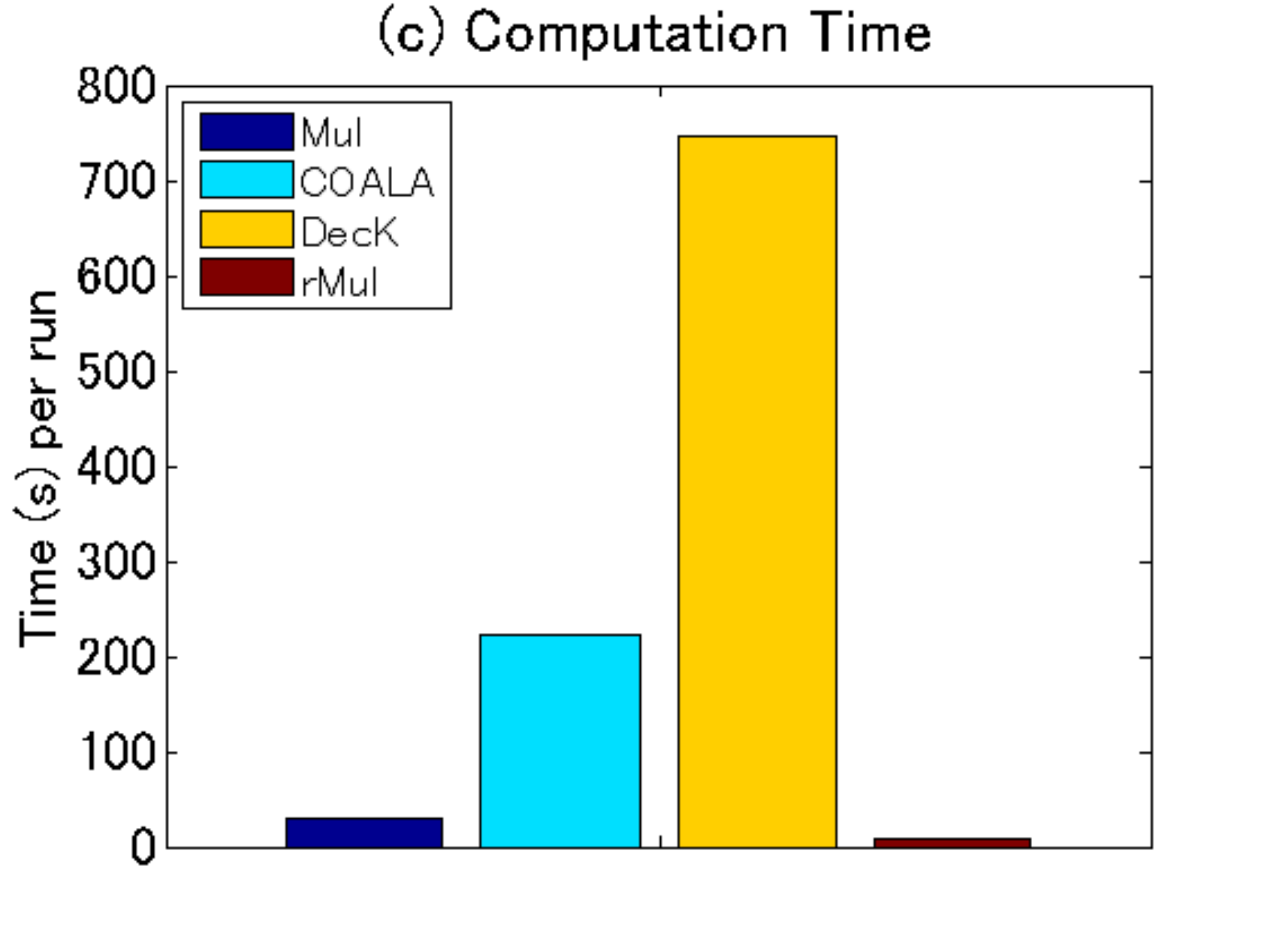} 
\caption{\it Performance on sample clusterings for the facial image data. Panel (a) for the subset 
(Data 1) of a single person (`an2i'). Performance on pose for four clustering methods, i.e., multiple co-clustering (Mul), COALA, decorrelated $K$-means (DecK), and restricted 
multiple clustering (rMul) are evaluated by adjusted Rand Index of sample clustering solutions. Panel (b) for the subset (Data 2) of two persons (`an2i' and `at33'). Performance is evaluated on useid and pose. Note that to match true and yielded views, we evaluated the maximum value of adjusted Rand index between the true sample
clustering in question and the yielded sample cluster solutions. The number of initializations is 500 for the multiple co-clustering, decorrelated $K$-means and the restricted multiple clustering. Panel (c) for computation time (seconds) per single run of each clustering method. 
}  
\label{imageres}
\end{figure}

Our multiple co-clustering method yielded nine sample clusterings (i.e., nine views), one of which is closely related to \textit{pose} with an adjusted Rand Index of 0.84 (Figure~\ref{imageres}a, $p<$0.001 by permutation test, and Table~\ref{clustermatrixface} for the contingency table between true clusters and resultant sample clusters). Our method outperforms COALA and decorrelated $K$-means methods (the performance of both methods is similar), and performs slightly better than the restricted multiple clustering method.

\begin{figure}[t]
\centering
\includegraphics[scale=0.22, trim=0mm 0mm 0mm 0mm]{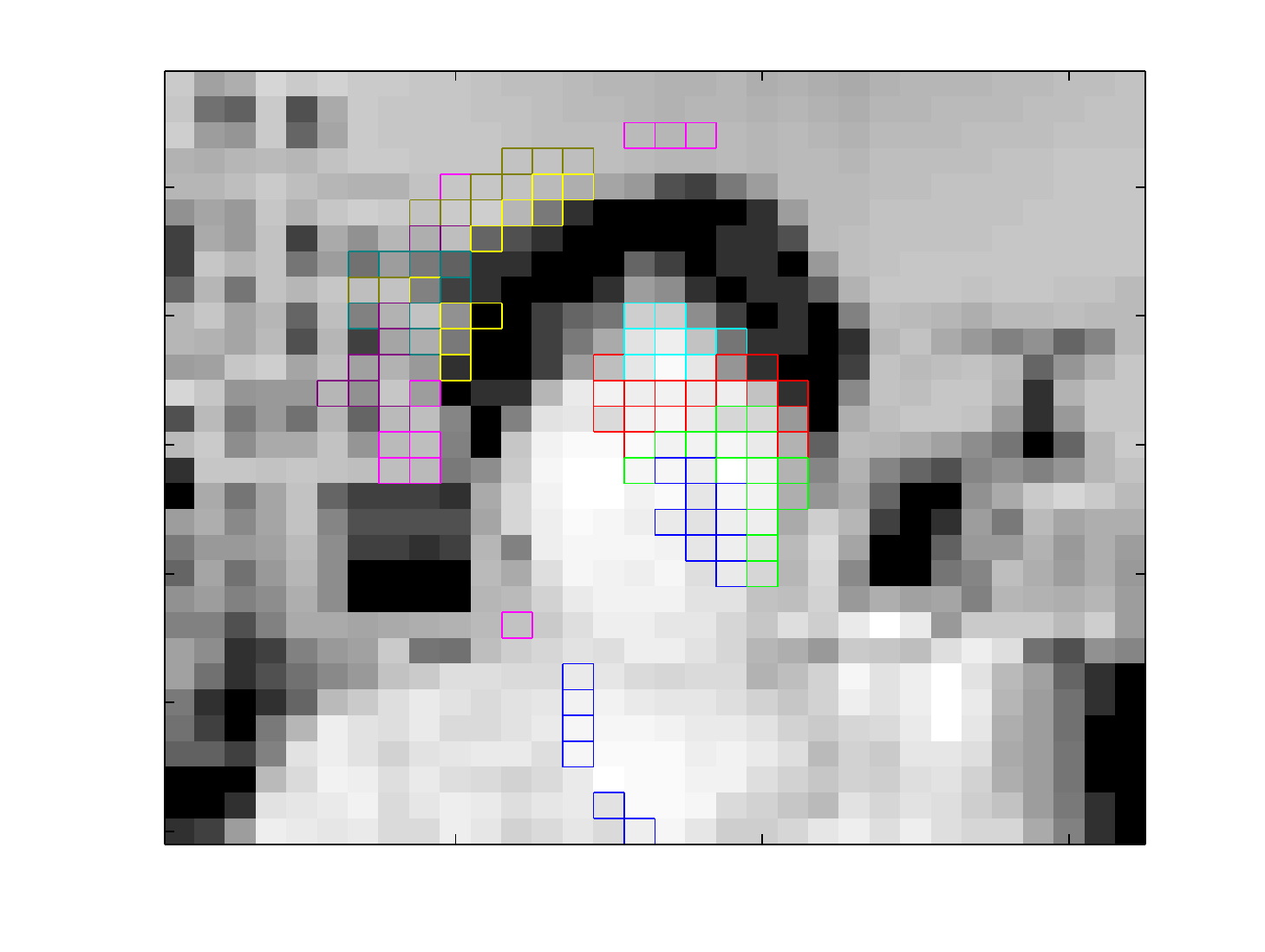} 
\includegraphics[scale=0.22, trim=0mm 0mm 0mm 0mm]{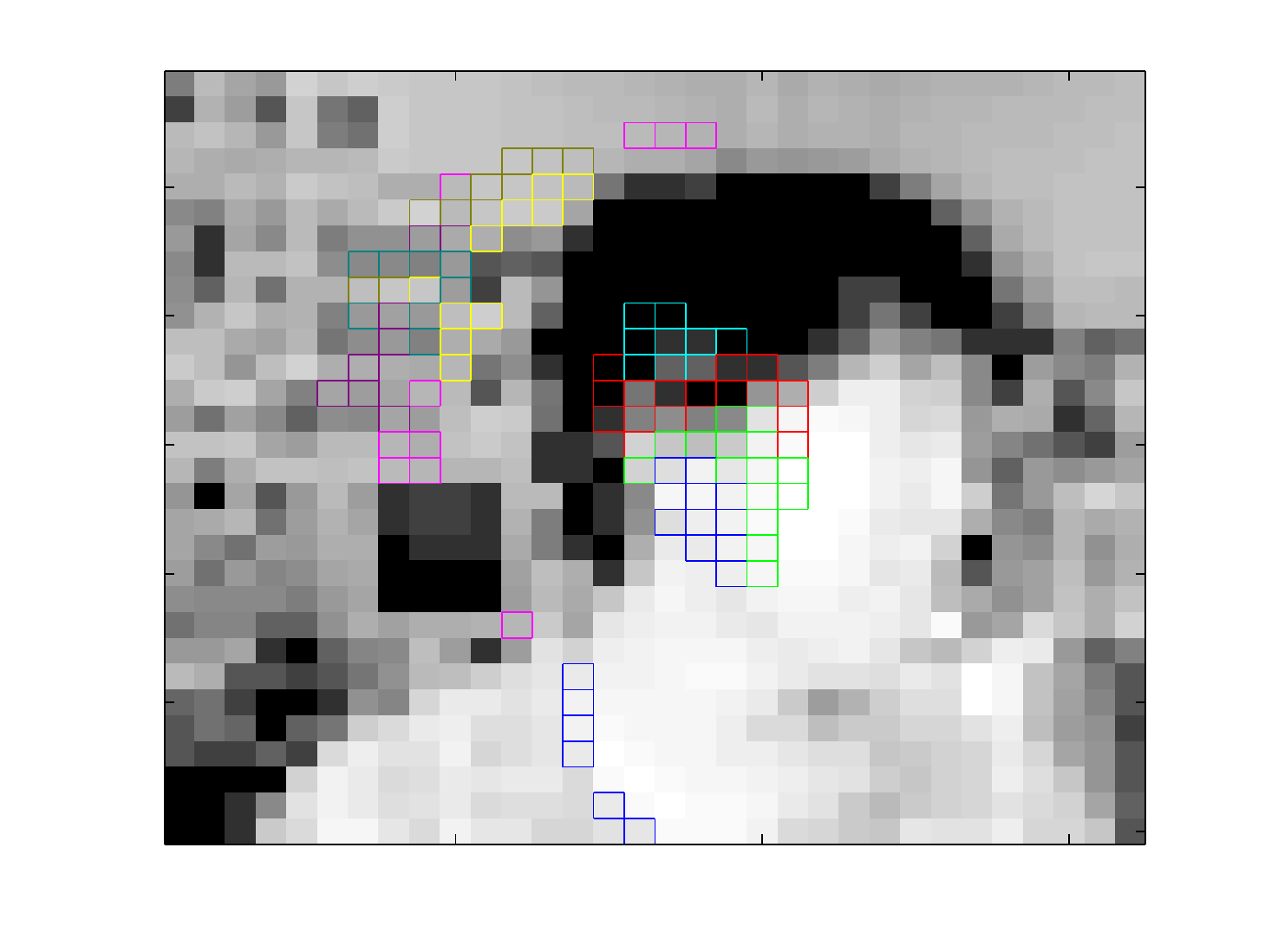} 
\includegraphics[scale=0.22, trim=0mm 0mm 0mm 0mm]{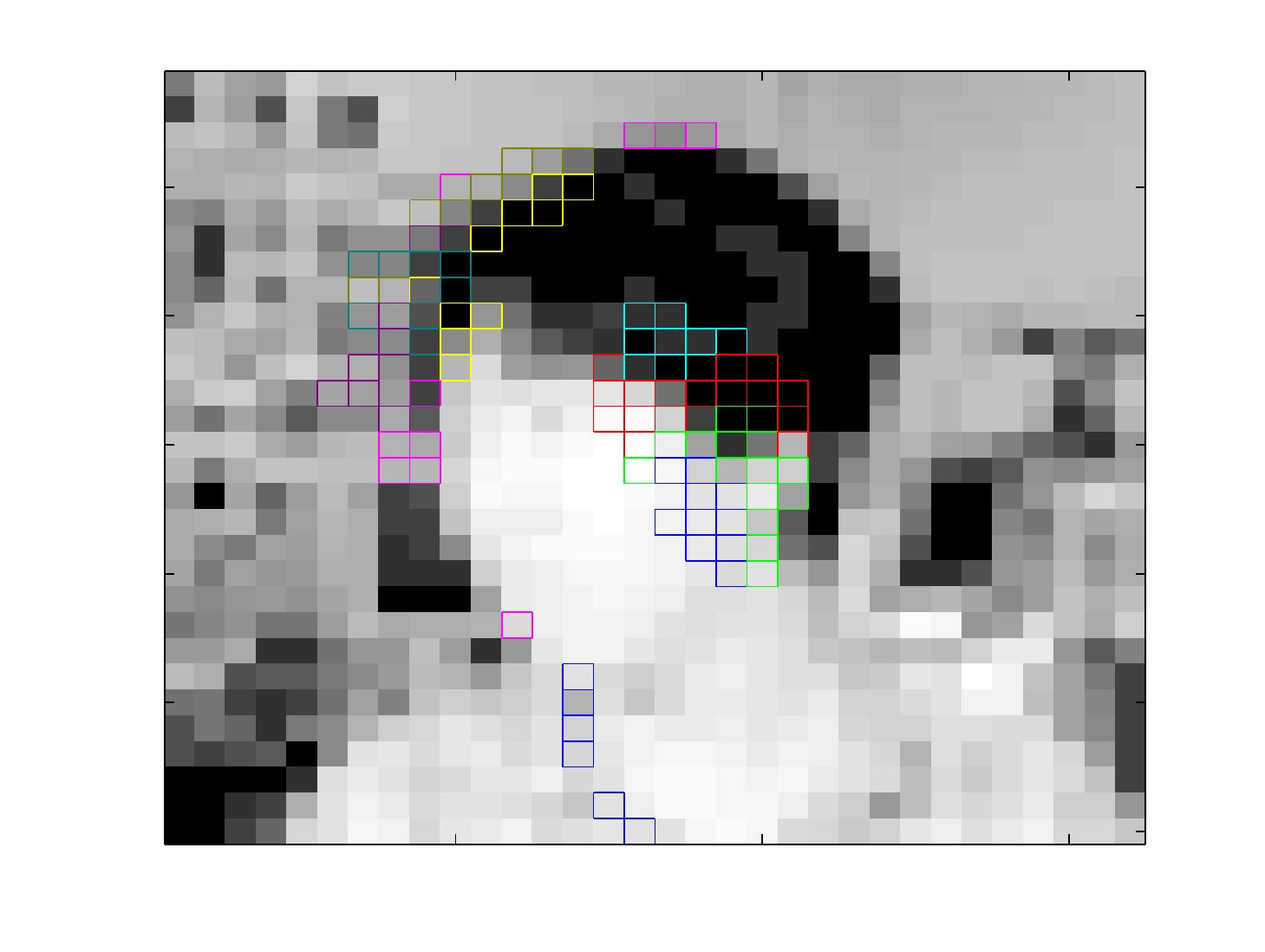} %
\includegraphics[scale=0.22, trim=0mm 0mm 0mm 0mm]{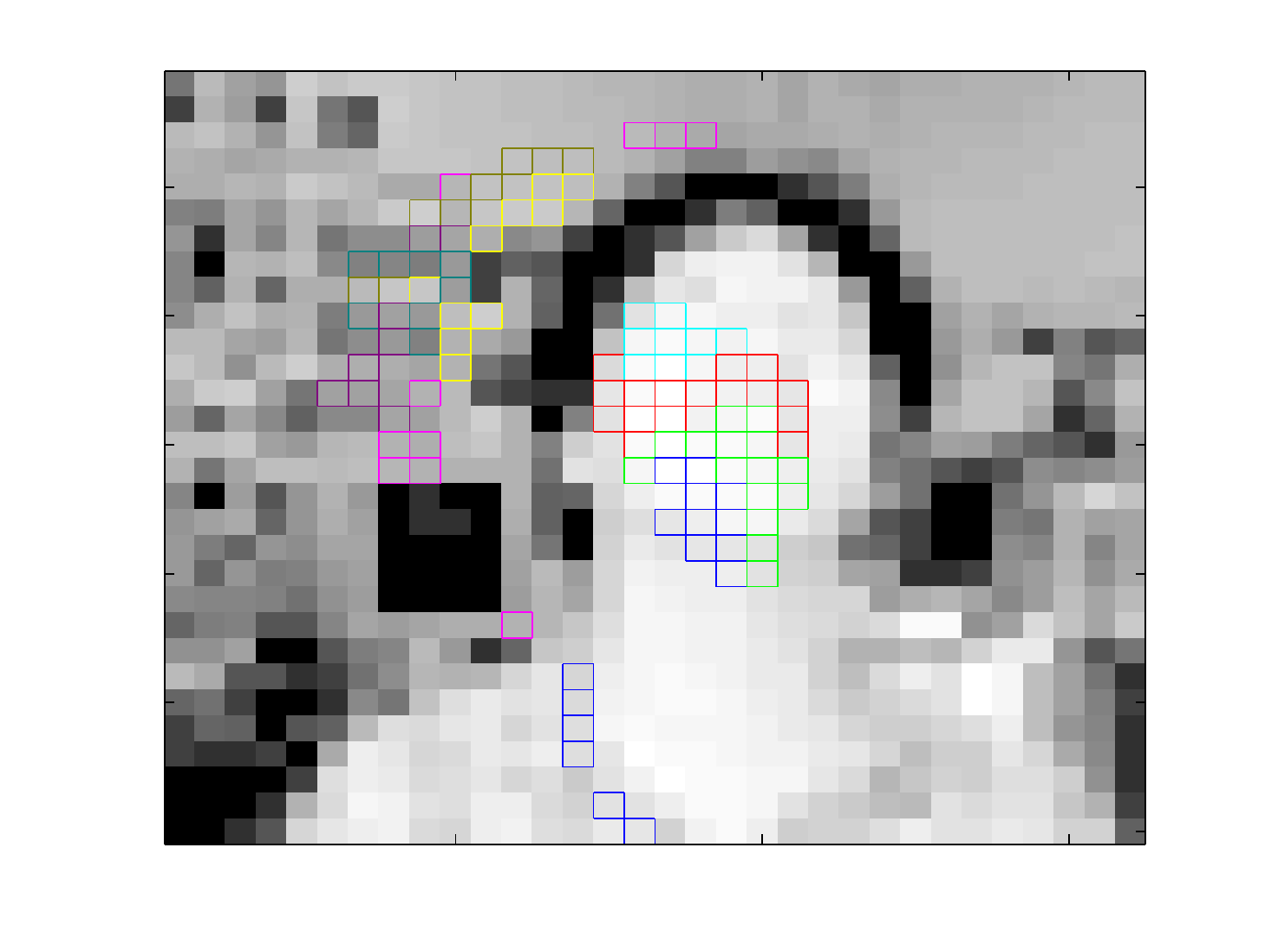} %
\caption{\it Selected features by our multiple co-clustering method for person `an2i' in the facial image data: Pixels surrounded by color boxes are the selected features that yielded the relevant sample clustering to pose. Color denotes a particular feature cluster. }
\label{image2}
\end{figure}

Next, we analyze features that are relevant to the sample clustering based upon  \textit{pose}. Note that our multiple co-clustering method yields information about features relevant to a particular sample clustering solution in an explicit manner while COALA and decorrelated $K$-means do not. Our method yielded the pixels (features) relevant to the cluster assignment, concentrating around subregions in the right part of head and the left part of face (Figure~\ref{image2}). This allows us to conclude that these subregions are very sensitive to different poses.

\begin{table}[!h]
\caption{\it Results of sample clustering 
for data 1 of the facial image data: Contingency table of the true labels (pose) and yielded clusters
of multiple co-clustering (Mul), COALA, decorrelated K-means (DecK), and restricted multiple (rMul) method from (a) to (d). T1, T2, T3 and T4 are true classes of pose  (straight, left, right and up); C1, C2, C3, C4 and C5 are yielded clusters 
for each method.}
\vspace{3mm}
\centering
\input{clustermatMulface.txt} \input{clustermatCOALAface.txt} \input{clustermatDecKface.txt} \\
\flushleft
\hspace{12mm}
\input{clustermatRestface.txt} \\
\label{clustermatrixface}
\end{table}

\begin{table}[!h]
\caption{\it Results for data 2 of the facial image data: Contingency table of the true labels (useid) and yielded clustering 
of the multiple co-clustering (Mul), COALA, decorrelated K-means (DecK), and restricted multiple (rMul) method from (a) to (d). T1 and T2 are true classifications (an2i, at33); C1, C2, C3 and C4 are yielded clusters.}
\vspace{3mm}
\centering
\input{clustermatMulface64useid.txt} \input{clustermatCOALAface64useid.txt} \input{clustermatDecKface64useid.txt} 
\input{clustermatRestface64useid.txt} \\
\label{clustermatrixface64useid}
\end{table}

\begin{table}[!h]
\caption{\it Results of sampling-clustering for data 2 of the facial image data: Contingency table of the true labels (pose) and yielded clustering 
of the multiple co-clustering (Mul), COALA, decorrelated K-means (DecK), and restricted multiple (rMul) method from (a) to (d). T1, T2, T3 and T4 are true classes of pose (straight, left, right and up); C1, ..., C7 are yielded results for each method.}
\vspace{3mm}
\centering
\input{clustermatMulface64.txt} \input{clustermatCOALAface64.txt} \input{clustermatDecKface64.txt} \\
\flushleft
\hspace{12mm}
\input{clustermatRestface64.txt} \\
\label{clustermatrixface64}
\end{table}

\subsubsection{Results: Data 2}
Our multiple co-clustering method yielded ten sample clusterings, three of which were closely related to \textit{useid}  (identification of person) and \textit{pose}  with adjusted Rand Indices of 0.82 ($p <$0.001) and 0.26 ($p <$0.001), respectively (Figure~\ref{imageres}b, and Tables~\ref{clustermatrixface64useid} and~\ref{clustermatrixface64} for the contingency tables for \textit{useid} and \textit{pose}, respectively). To compare with COALA, our multiple co-clustering method performed a bit poorly for detecting \textit{useid}, while it performed better for \textit{pose}. On the other hand, the performance of our method is comparable to the decorrelated $K$-means method. Further, it performed slightly better than the restricted multiple clustering method. 

\begin{figure}[!t]
\centering
\includegraphics[scale=0.22, trim=0mm 0mm 0mm 0mm]{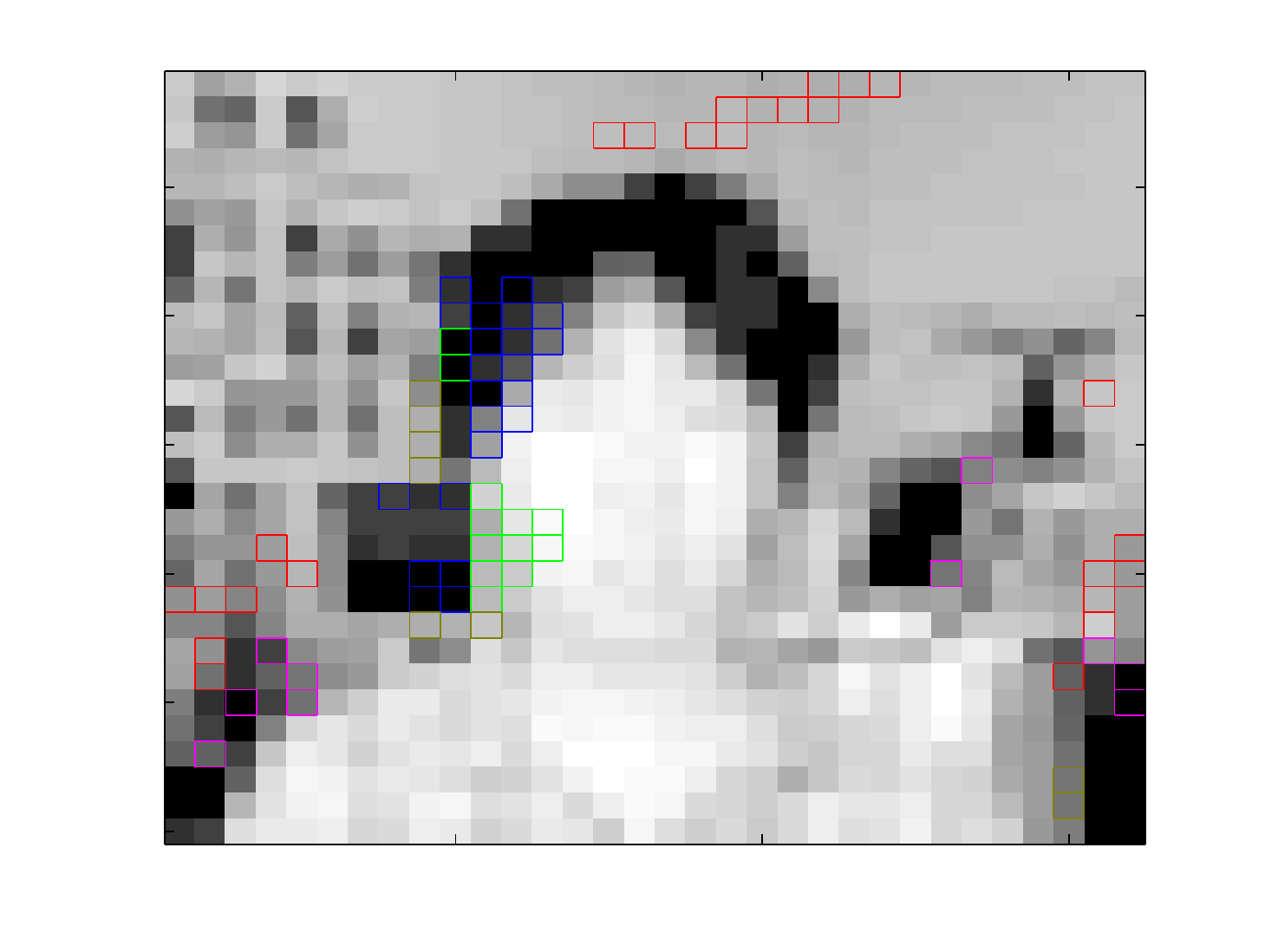} 
\includegraphics[scale=0.22, trim=0mm 0mm 0mm 0mm]{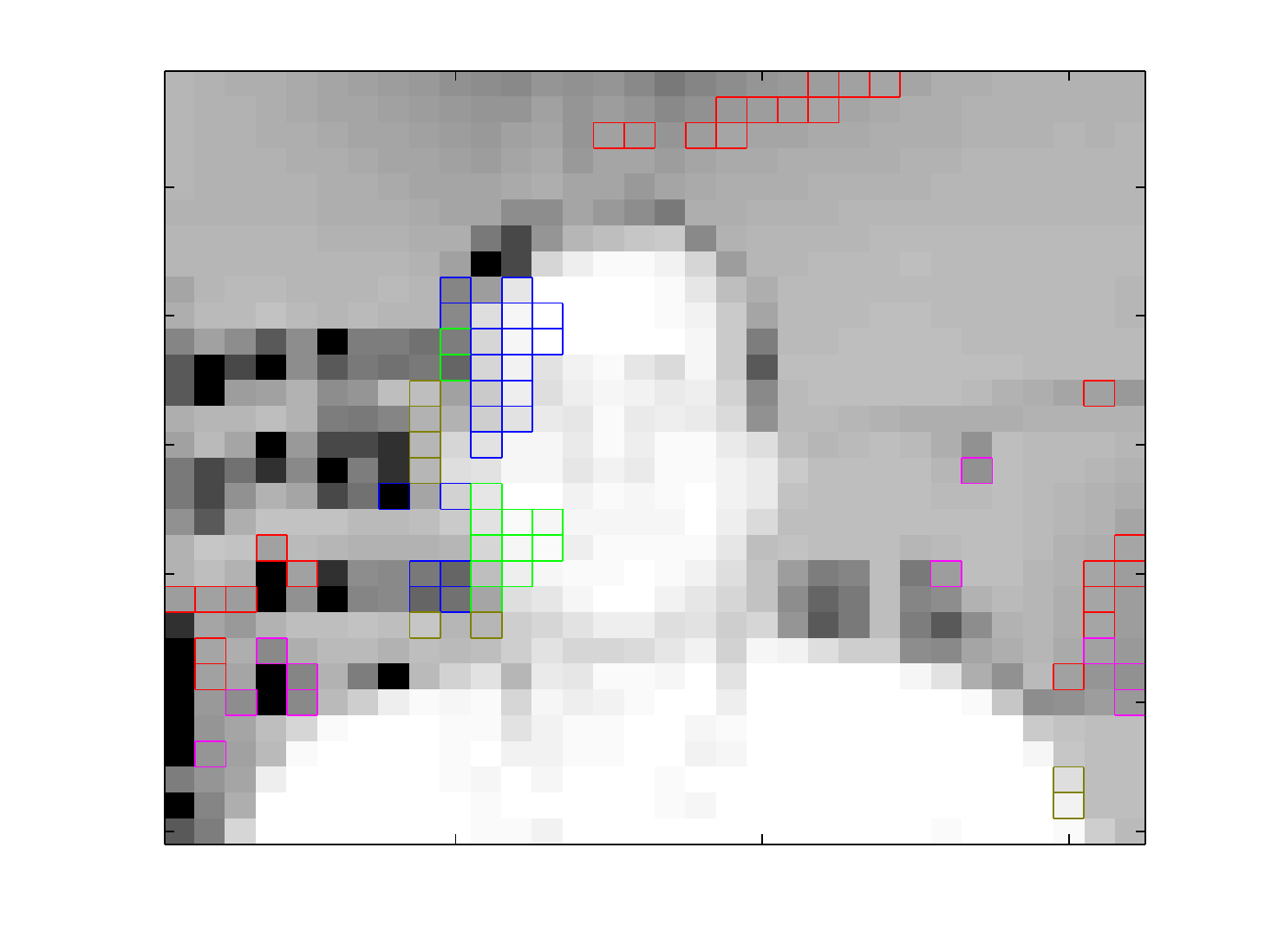} 
\includegraphics[scale=0.22, trim=0mm 0mm 0mm 0mm]{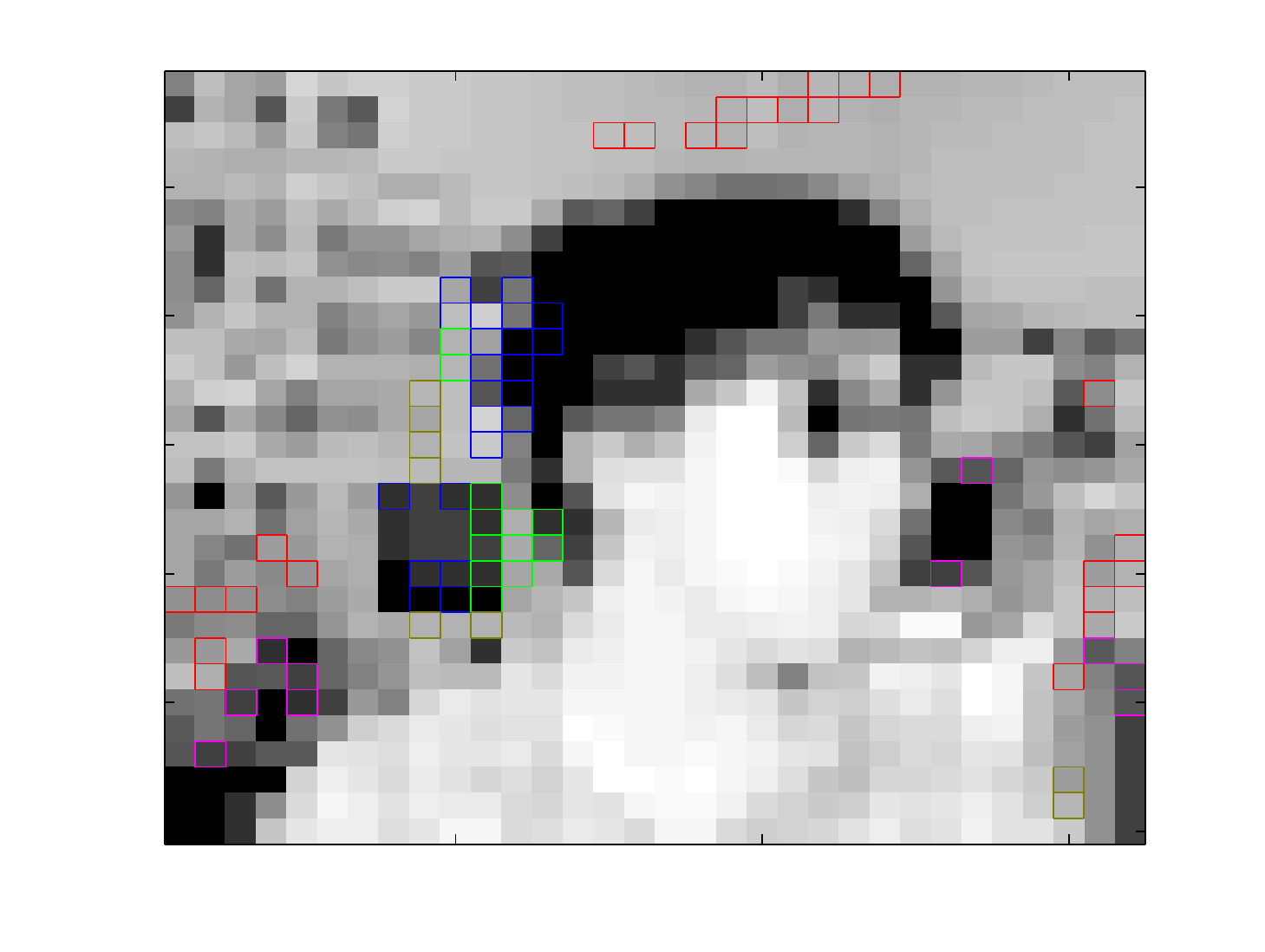} %
\includegraphics[scale=0.22, trim=0mm 0mm 0mm 0mm]{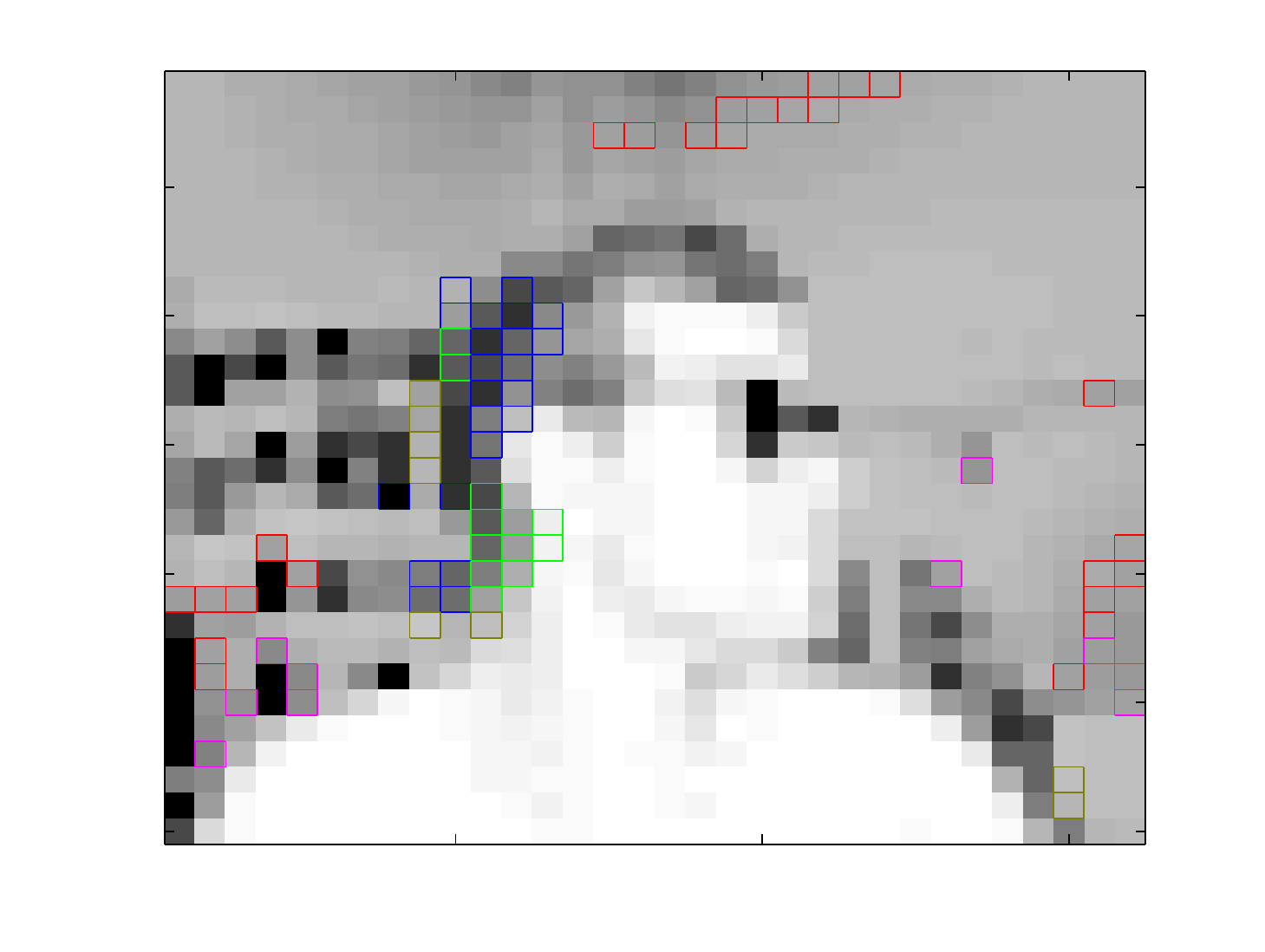} %
\caption{\it Samples from image datasets for person `an2i' and `at33': Pixels surrounded by color boxes are selected features that yielded relevant sample clustering to useid in data2. Image configurations are (`an2i', non sunglass, straight), (`at33', non sunglass, straight),
`an2i', sunglass, left), and (`at33', sunglass, left), respectively. Expression is neutral for
all samples. In these examples, the multiple clustering method correctly identified these persons. 
}  
\label{image3}
\end{figure}

The most relevant pixels for \textit{useid} concentrate near the right part of face, and the background (Figure~\ref{image3}). This can be interpreted to mean that the difference in hair style may be an important factor to distinguish between these two persons. In addition, an apparent difference in their rooms (background) also serves as a discriminating factor.

\subsection{Cardiac Arrhythmia data}
Next, we apply our multiple co-clustering method to Cardiac Arrhythmia data (UCI KDD repository). Unlike the facial image data in the previous section, this dataset does not necessarily have multiple sample clustering structures (indeed, such information is not available). However, the multiple co-clustering method should be able to automatically select relevant features.

The original dataset consisted of 452 samples (subjects) and 279 features that comprise various cardiac measurements and personal information such as sex, age, height, and weight (See more detail in \cite{guvenir1997supervised}). Some of these features are numerical (206 features) and others are categorical (73 features). Further, there are a number of missing entries in this dataset. Beside these features, a classification label is available, which classifies the subjects into one of 16 types of arrhythmia. For simplicity, we focus only on three types of arrhythmia of similar sample size: \textit{Old Anterior Myocardial infarction} (sample size 15), \textit{Old Inferior Myocardial Infarction} (15) and \textit{Sinus Tachycardia} (13). The objective in this section is to examine recovery performance among these three types of arrhythmia.

Application of COALA and decorrelated $K$-means methods to this dataset is problematic because these methods do not allow for categorical features nor missing entries. Hence, we use the following heuristic procedure to pre-process the data: Re-code a binary categorical feature using a numerical feature (taking values 0 or 1); replace missing entries with mean values of features. Recall that these problems do not arise with our multiple co-clustering method.

\subsubsection*{Results}
\begin{figure}[!h]
\centering
\includegraphics[scale=0.4]{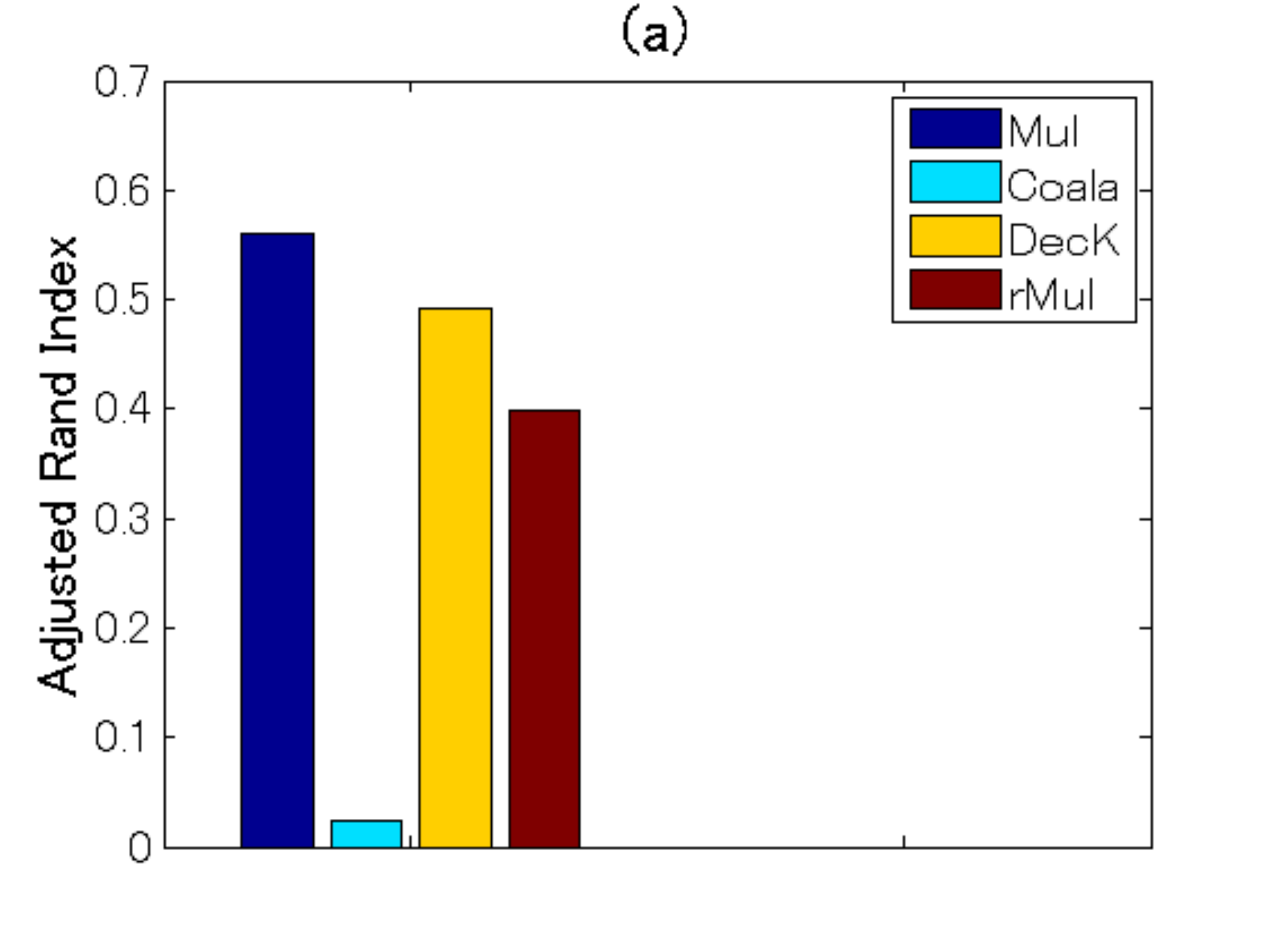} 
\caption{\it  Results of multiple clustering for the cardiac arrhythmia data: 
Comparison of performance on subject clustering in terms of adjusted Rand Index
among multiple co-clustering, COALA,
decorrelated $K$-means and restricted multiple clustering methods}  
\label{ArrhyARI}
\end{figure}

Our multiple co-clustering method yielded nine sample clusterings (i.e., nine views). The maximum adjusted Rand Index between the true labels and resultant clusters is 0.56 ($p<$0.001, Figure~\ref{ArrhyARI}). On the other hand, the maximum Rand Index for COALA, decorated K-means and restricted multiple clustering methods are 0.02, 0.49 and 0.39, respectively.

\begin{table}[!h]
\caption{\it Results of sample clustering for the cardiac arrhythmia data: Contingency table of the true labels and yielded clustering of the multiple co-clustering (Mul), decorrelated K-means (DecK), and restricted multiple (rMul) method from (a) to (d). T1, T2, and T3 are true classes of arrhythmia (Old Anterior Myocardial Infarction, Old Inferior Myocardial infarction, and Sinus Bradycardy, respectively); C1, C2, C3 and C4 are yielded results for each method.}
\vspace{3mm}
\centering
\input{Matrix1.tex} \input{MatrixCOALA2.tex}  \input{MatrixK1.tex} \input{MatrixRest6.tex}\\
\label{clustermatrixCardiac}
\end{table}

Further, we examine subject clustering results more in detail. For our multiple co-clustering method, the subject clusters C1, C2, and C3 distinguish the three symptoms well (corresponding to T2, T1, and T3, respectively, Table~\ref{clustermatrixCardiac}). On the other hand, such a distinction is totally or partially ambiguous for the other methods. In the case of COALA, clustering results seem to be degenerate, yielding two tiny clusters (C2 and C3). For decorrelated $K$-means, the distinction among T1, T2, and T3 is partially ambiguous. There is a clear correspondence between C1 and T1, but C2 is a tiny cluster, and C3 does not distinguish between T2 and T3. A similar observation is made for the restricted multiple clustering method.

\subsection{Depression data}\label{applicationreal}
Lastly, we apply our multiple co-clustering method to depression data, which consists of clinical questionnaires and bio-markers for healthy and depressive subjects. The objective here is to explore ways of analyzing the results from our multiple co-clustering method in a real situation where the true subject-cluster structures are unknown. The depression data comprise 125 subjects (66 healthy and 59 depressive) and 243 features (Table~\ref{association} in Appendix~\ref{listfeatures}) that were collected at a collaborating university. Among these features, there are 129 numerical (e.g., age, severity scores of psychiatric disorders) and 114 categorical features (e.g., sex, genotype) with a number of missing entries. Importantly, these data were collected from subjects in three different phases. The first phase was when depressive subjects visited a hospital for the first time. The second phase was 6 weeks after subjects started medical treatment. The third phase was 6 months after the onset of the treatment. For healthy subjects, relevant data for the second and the third phases were not available (dealt as missing entries in the data matrix). To distinguish between these phase differences, we denote features in the second and the third phases with endings of $6w$ and $6m$, respectively. Further, we did not include the label of health/depression status for clustering. We used it only for interpretation of results. We assumed that numerical features follow mixtures of Gaussian distributions in our model. To pre-process numerical features, we standardized each of them using means and standard deviations of available (i.e., nonmissing) entries.

\begin{figure}[h]
\centering
\includegraphics[scale=0.35, trim=0mm 0mm 0mm 0mm]{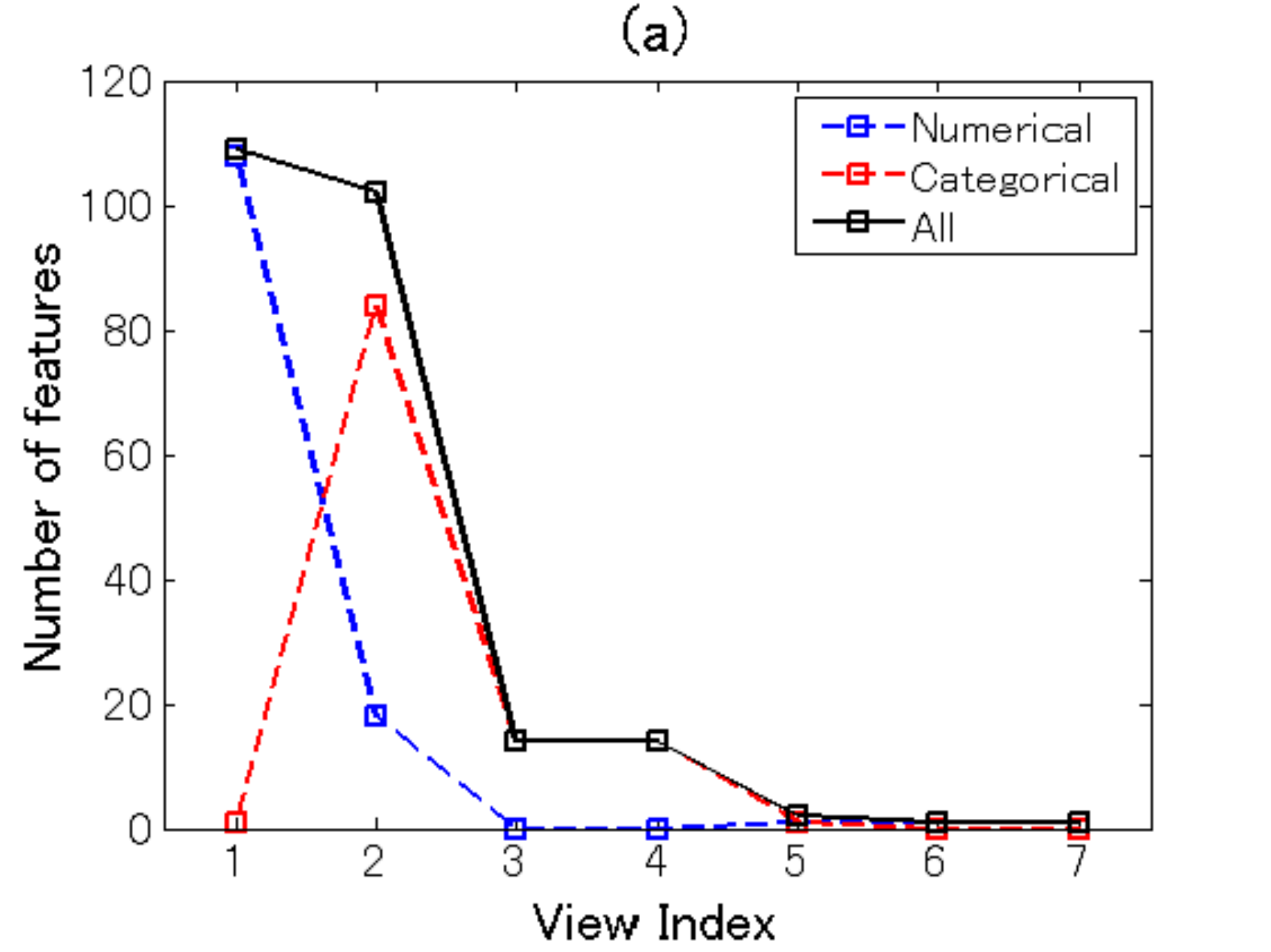} 
\includegraphics[scale=0.35, trim=0mm 0mm 0mm 0mm]{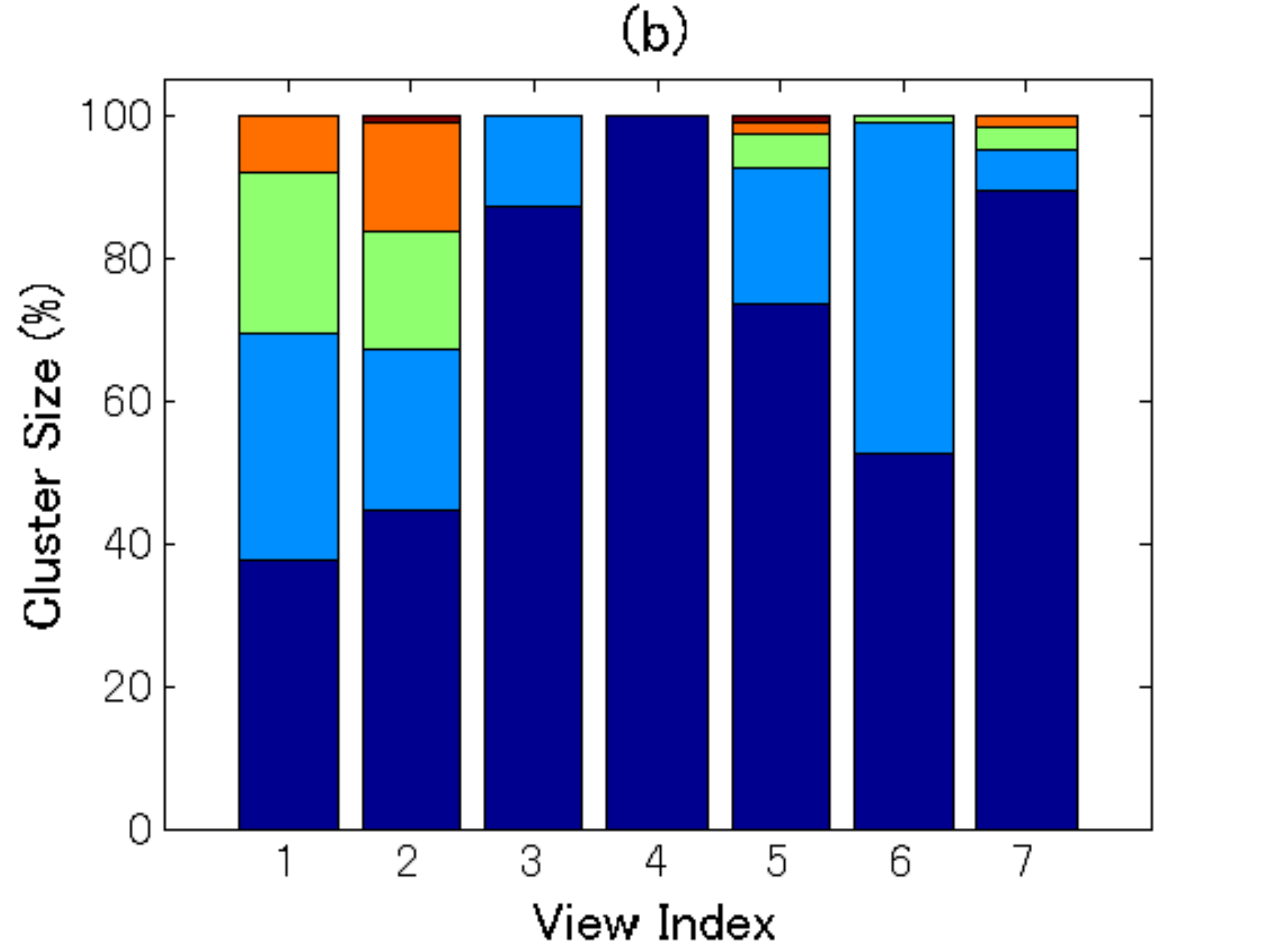} 
\caption{\it Results of the multiple co-clustering method for clinical data of depression. Panel (a): Number of features (in black) in each view with numerical features in blue and categorical features in red. Panel (b):
cluster size (percentage of subjects) for subject clusters in each view. 
}  
\label{depClusterResults}
\end{figure}

\subsubsection*{Results}
\begin{figure}[!h]
\centering
\includegraphics[scale=0.41, trim=0mm 0mm 0mm 0mm]{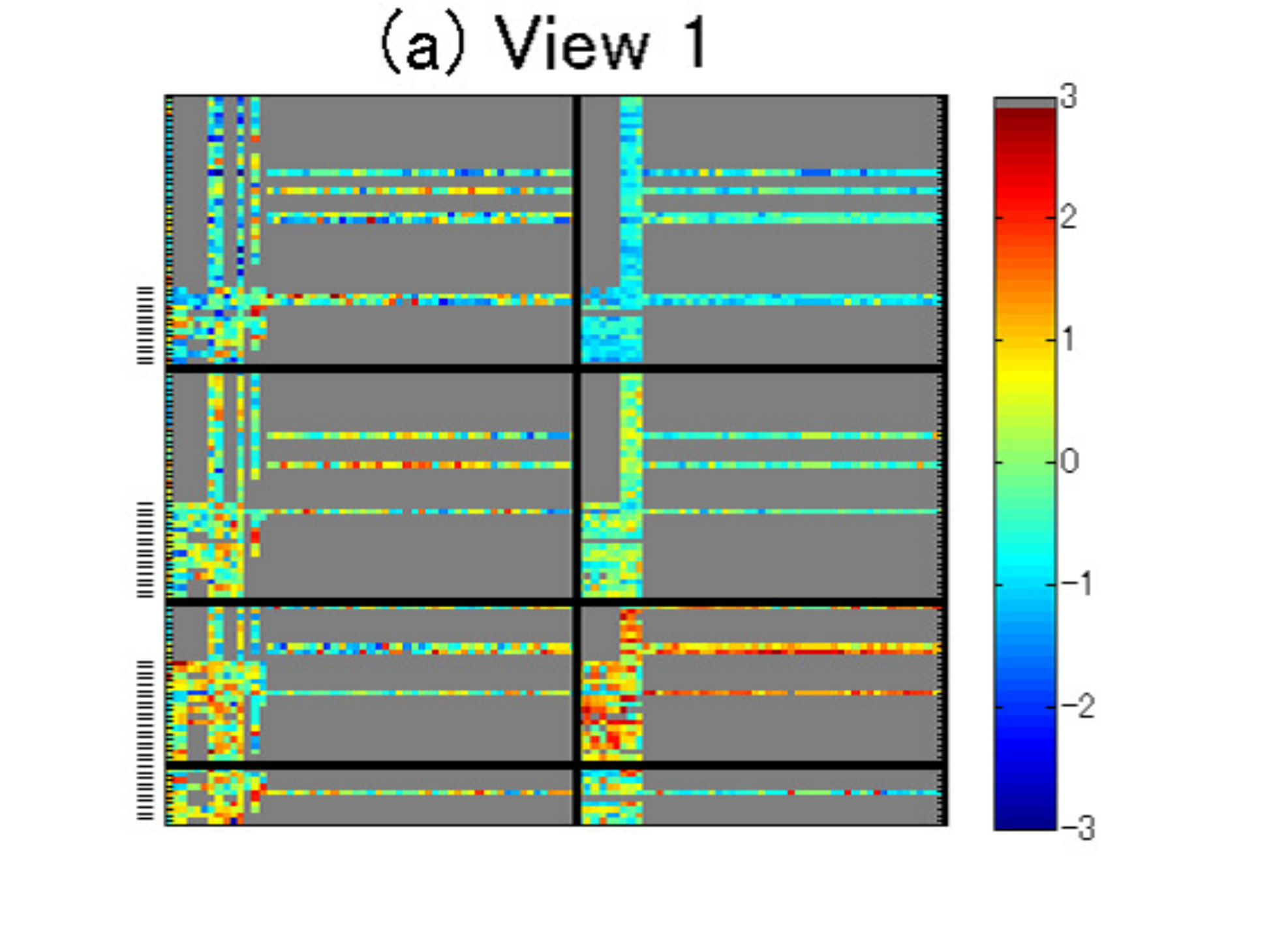} 
\includegraphics[scale=0.41, trim=0mm 0mm 0mm 0mm]{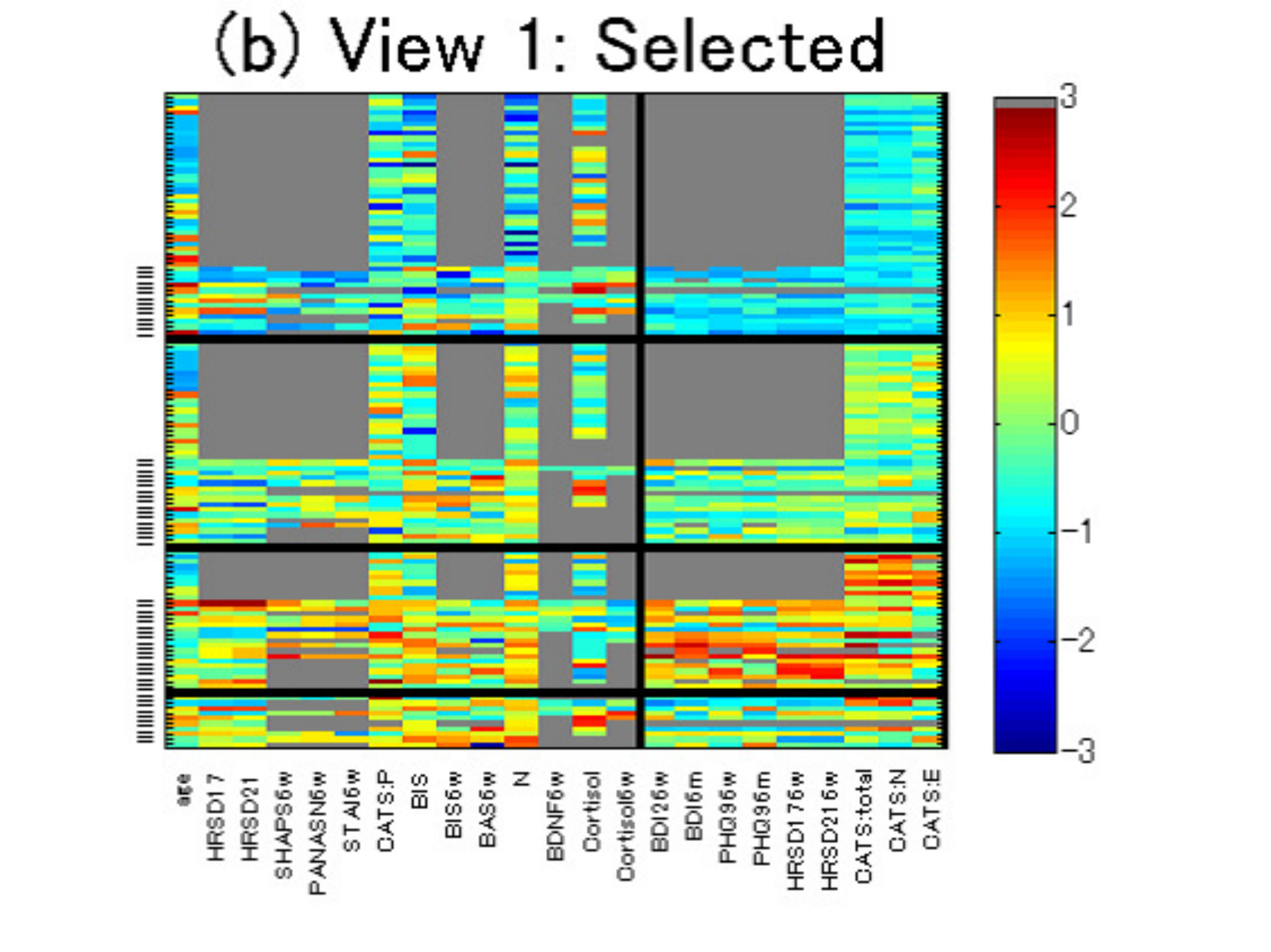} 
\includegraphics[scale=0.41, trim=0mm 0mm 0mm 0mm]{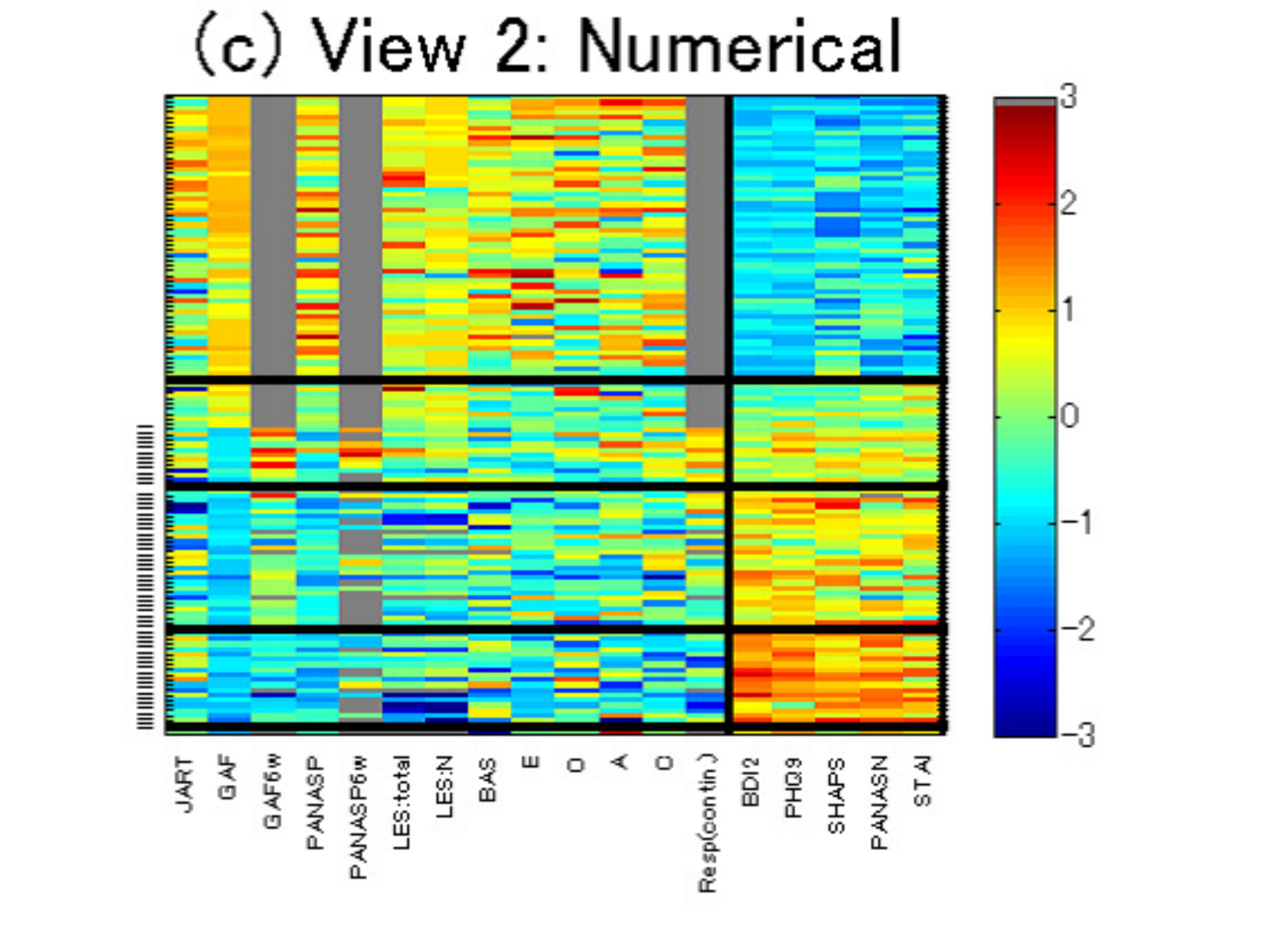} 
\includegraphics[scale=0.41, trim=0mm 0mm 0mm 0mm]{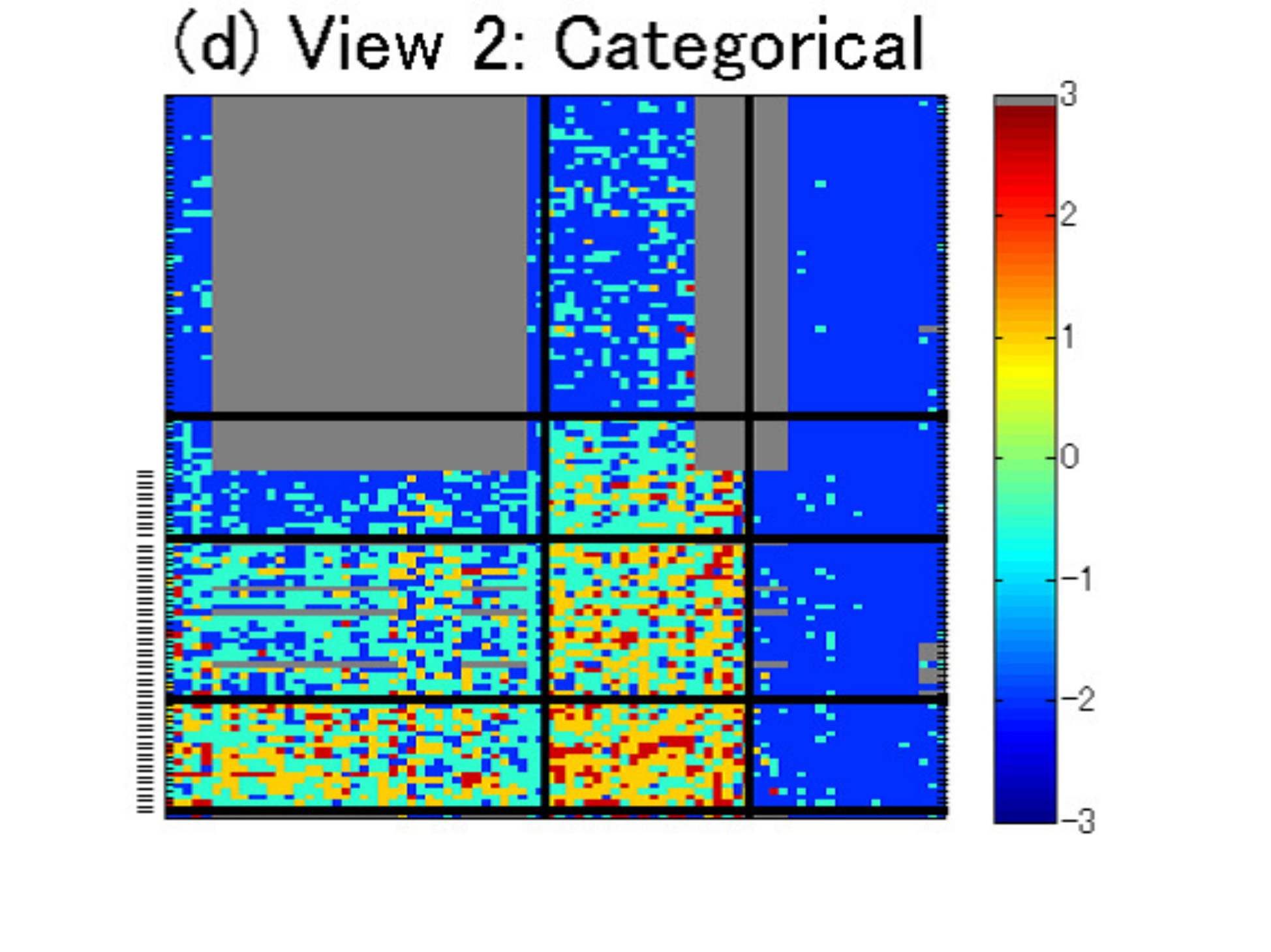} 
\caption{\it Visualizations of views yielded by our multiple co-clustering method.
Panels (a)-(b): Heatmaps of views 1. The
x-axis denotes numerical features, and the y-axis denotes subjects. A depressive subject is indicated by 
a hyphen in left. The subject clusters are sorted in the order of cluster size. Panel (b) is a copy of panel (a) after removing methylation related features (those having a large number of missing entries).  
Panels (c)-(d): Heatmaps of views 2. Panel (c) 
contains numerical features while panel (d) contains categorical ones. 
The subject clusters are sorted in the descending order of the proportion of depressive subjects. For these panels, the subjects within a subject cluster are sorted in the order of healthy and depressive subjects. On the other hand, feature clusters are sorted in the order of feature clusters  in the order of feature cluster size. 
Note that for categorical features the color is arbitrary and that missing entries are in gray.
}  
\label{depClusterResults2}
\end{figure}

\begin{figure}[!h]
\centering
\includegraphics[scale=0.41, trim=0mm 0mm 0mm 0mm]{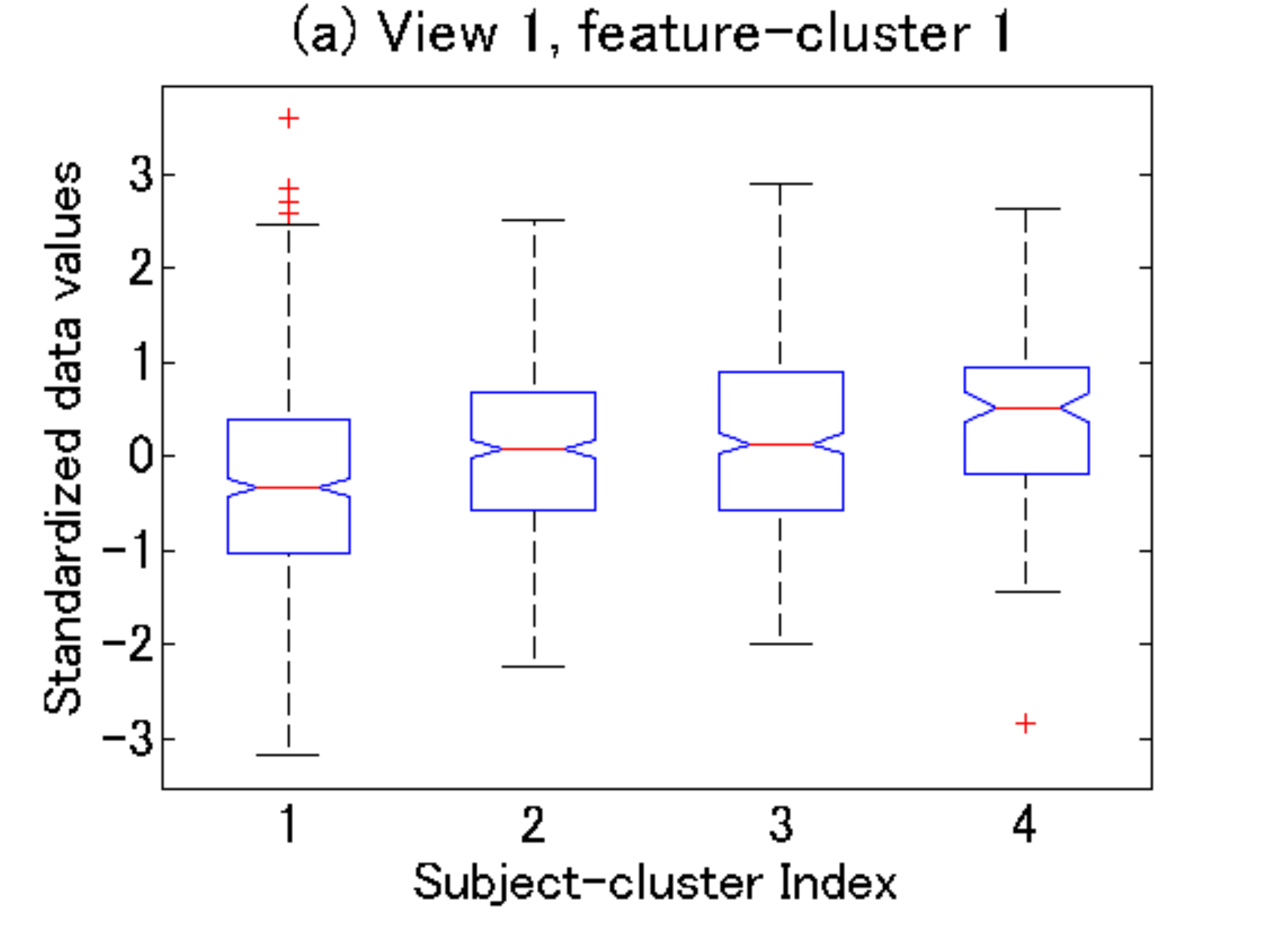} 
\includegraphics[scale=0.41, trim=0mm 0mm 0mm 0mm]{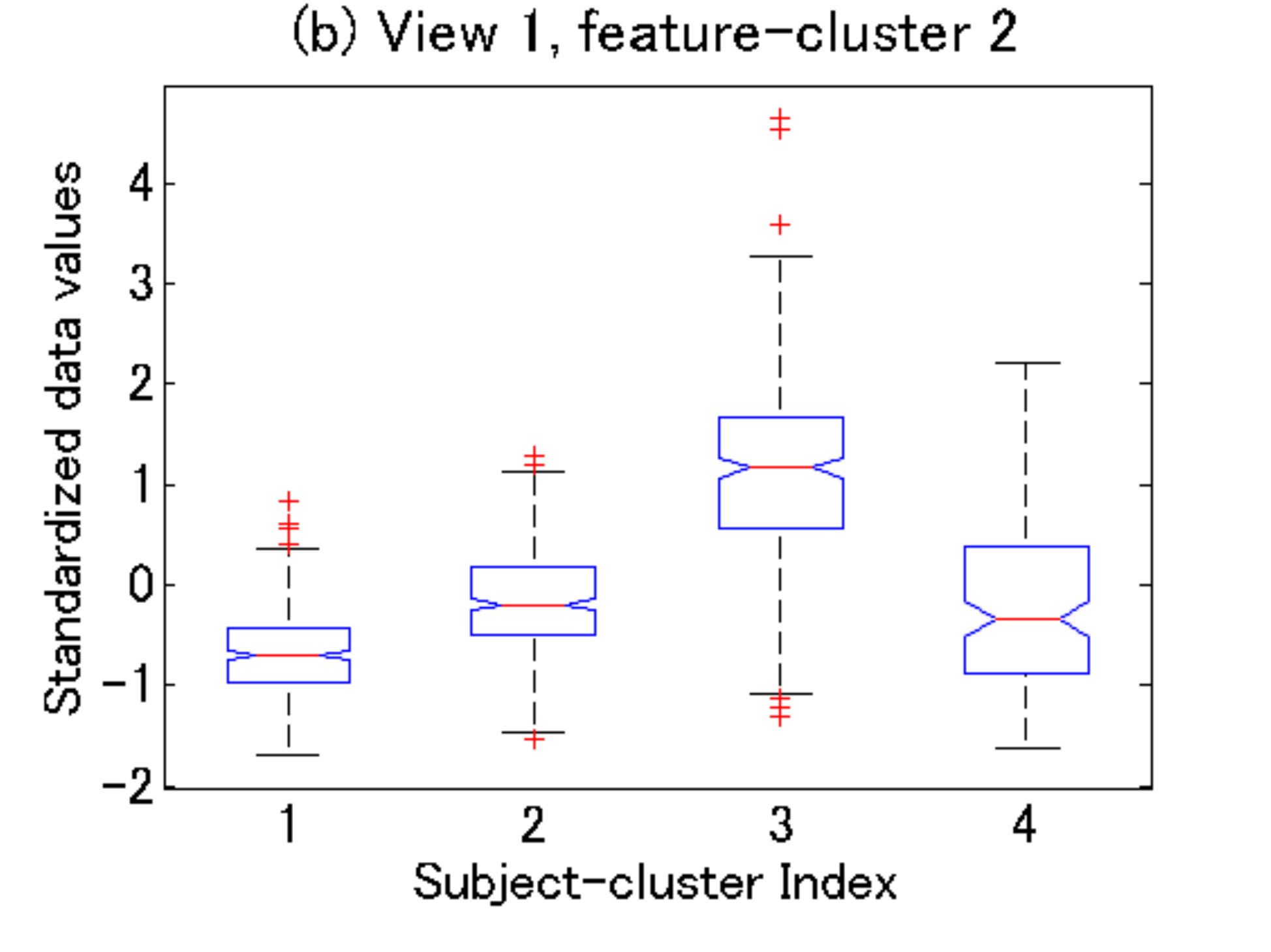} 
\caption{\it Distributions of the standardized data within feature clusters 1 (Panel (a)) and 2 (Panel (b)). X-axis
denotes subject cluster index. All relevant entries except for missing ones are accommodated in each box. 
}  
\label{boxplotdis}
\end{figure}

Our multiple co-clustering method yielded seven views. The majority of features are allocated to two views (view 1 and view 2 in Figure~\ref{depClusterResults}a). The number of subject clusters ranges from one to five (Figure~\ref{depClusterResults}b). We analyze these cluster results more in detail, focusing on view 1 and view 2. View 1 has two feature clusters for numerical features, in which the majority of features are related to DNA methylation of CpG sites of the trkb and htr2c genes with a number of missing entries (Figure~\ref{depClusterResults2}a). For better visualization of this view, we remove methylation-related features (Figure~\ref{depClusterResults2}b). Among these two (numerical) feature clusters, feature cluster 1 does not discriminate well between the yielded subject clusters (Figure\ref{boxplotdis}a), while feature cluster 2 does well (Figure~\ref{boxplotdis}b). Hence, subject clustering in this views is largely characterized by features in feature cluster 2 (BDI26w, BDI26m, PHQ96w, PHQ96m, HRSD176w, HRSD216w, CATS:total, CATS:N, and CATS:E). The first six features are related to psychiatric disorder scores at six weeks (features ending with -$6w$) and six months (features ending with -$6m$) after the onset of depression treatment. Hence, we can interpret this to reflect treatment effects. On the other hand, CATS:total, CATS:N, and CATS:E are related to abusive experiences in the subject's childhood. Hence, these features are available before the onset of treatment. These data attributes suggest that it is possible to predict treatment effect by using features related to child abuse experiences. In particular, the distribution pattern in subject cluster 3 (Figure~\ref{depClusterResults2}b) is remarkably different from those in the remaining subject clusters (Figure~\ref{boxplotdis}b).

In view 2, healthy and depressive subjects are well separated. The first subject cluster is for healthy subjects. The second is intermediate, and the third and fourth are for depressive subjects (Figure~\ref{depClusterResults2}c). Relevant numerical features are: JART, GAF, GAF6w, PANASP, PANASP6w, LES:total, LES:N, BAS, E, O, A, C, and Rep. in feature cluster 1 and BDI2, PHQ9, SHAPS, PANASN, and STAI in feature cluster 2. This result is quite reasonable, because the majority of these features are scores from clinical questionnaires that evaluate depressive disorders either negatively (smaller values in feature cluster 1) or positively (larger values in feature cluster 2).

\subsection{Comparison of time complexity} \label{complexity}
Finally, we briefly discuss complexity of the clustering methods. Except for COALA, we need to run clustering methods (i.e., multiple co-clustering, decorrelated $K$-means, and restricted multiple clustering) with a number of random initializations for their parameters, and choose the optimal solution. Hence, computation time depends on the number of initializations. To compare complexity of computation, we make several assumptions. First, we focus on a single run of each method. Second, we assume the same number of iterations for convergence. Third, we assume that the numbers of views and clusters are fixed. In such a setting, time complexity for our multiple clustering method (as well as the restricted multiple clustering) is $O(nd)$, where $n$ and $d$ are the number of samples and the number of features, respectively (Section \ref{timecomplex}). This suggests that the complexity is just linear when either $n$ or $d$ is fixed. On the other hand, the complexities of COALA and decorrelated K-means are $O(n^2 \log n + n^2d)$ and $O(nd + d^3)$, respectively, based on their typical algorithms \cite{bae2006coala, jain2008simultaneous}. These results imply that the complexity of our multiple co-clustering is generally less than those of COALA and decorrelated $K$-means, suggesting superior efficiency of the present method. Indeed, in the simulation of facial image data, our multiple clustering method requires less time per run than COALA and decorrelated $K$-means (Figure~\ref{imageres}c).

\section{Discussion}
We propose a novel method of multiple clustering in which each view comprises a co-clustering structure, and each cluster block fits a univariate distribution. Though our method assumes a somewhat complicated cluster structure (multiple views of co-clustering structures), it effectively detects multiple cluster solutions by clustering relevant features within a view, based on their distributional patterns. In contrast with our multiple co-clustering method, the restricted multiple clustering method is simple and straightforward for implementation. However, from a factor-analytical perspective, fitting a single distribution to all features in a view implies the dimensional reduction of that view by a single factor, which may be too restrictive for high-dimensional data. On the other hand, our method relaxes this constraint, allowing possible factors to be inferred in a data-driven approach. Practically, if there is prior knowledge that each view consists of a single factor, then we may use the restricted multiple clustering method. Otherwise, it is preferable to use our multiple co-clustering method, as demonstrated in both synthetic and real data applications above.

In comparisons with COALA and decorrelated $K$-means methods, our multiple co-clustering method outperforms other state-of-the-art methods using facial image and cardiac arrhythmia data. Beyond its better performance in sample clustering, our multiple co-clustering method has several advantages over other methods. It can infer the number of views/clusters. It is applicable to datasets comprising different types of features, and it can identify relevant features. Furthermore, our method is computationally efficient. The reason for this efficiency is the fitting of a univariate distribution to each cluster block. It is notable that despite using only a univariate distribution, our multiple co-clustering method can flexibly fit a dataset by adapting the number of views/clusters by means of a Dirichlet process.

Finally, it is worth noting that the multiple co-clustering method is not only useful to recover multiple cluster structures of data, but also a single-cluster structure. In the case of single clustering, our method works by selecting relevant features for possible sample clustering. This may be the main reason that our method performs better with the cardiac arrhythmia data than COALA and decorrelated $K$ means, which use the data without feature selection.

\section{Acknowledgement}
This study is the result of ``Integrated Research on Neuropsychiatric Disorders"
carried out under the Strategic Research Program for Brain Science by the Ministry of Education,
Culture, Sports, Science and Technology of Japan. We would like to thank Dr. Steve D. Aird at Okinawa Institute 
of Science and Technology Graduate University for his proof-reading of this article.


\begin{appendices}
\section{Observation models for Gaussian, Poisson and multinomial} \label{appenobs}
This is a supplementary material for Section~\ref{obsmodel} (Observation models) in the 
main manuscript, providing priors, the expectation of log-likelihood and the updating equations. 

\subsection{Gaussian distribution}
We denote univariate Gaussian density function as $\mbox{Gauss}(\cdot|\mu, \sigma^2)$ where $\mu$ and $\sigma^2$ are mean and variances. We assume conjugate priors for $\mu$ and $\sigma^2$ in each cluster block: 
\begin{eqnarray*}
s_{v, g, k} &\sim& \mbox{Ga}(\cdot|\gamma_0/2,\gamma_0\sigma_0^{2}/2)\\
\mu_{v, g, k} &\sim &\mbox{Gauss}(\cdot|\mu_0, (\lambda_0 s_{v, g, k})^{-1}),
\end{eqnarray*}
where $\mbox{Ga}(\cdot|a, b)$ denotes Gamma distribution with shape and rate parameters $(a, b)$.
In the present paper, we set $\sigma_0^{2}=10^{4}$, $\gamma_0=1$, and $\lambda_0=10^{-4}$ so that the prior distributions are nearly non-informative.
It can be shown that the variational approximation for the posterior 
$q_{\bb{\theta}^{(m)}}(\bb{\theta}^{(m)})$ is given by
\begin{eqnarray*}
&& \prod_{v=1}^V \prod_{g=1}^{G} \prod_{k=1}^K
 \mbox{Gauss}(\mu_{v, g, k}|\mu_{0, v, g, k}, (\lambda_{0, v, g, k}s_{0, v, g, k})^{-1}) \\
 &&~~~~~~~~~~\times \mbox{Ga}(s_{0, v, g, k} | \gamma_{0, v, g, k}/2, \gamma_{0, v, g, k} \sigma_{0, v, g, k}^{2}/2),
\end{eqnarray*}
where the hyperparameters are updated by
\begin{eqnarray*}
 \lambda_{0, v, g, k} &=& \lambda_0 + \sum_{j=1}^{d^{(m)}}\sum_{i=1}^n \tau^{(m)}_{j, v, g}\eta_{i, v, k}\\
 \mu_{0, v, g, k} &=& \frac{1}{\lambda_{0, v, g, k}} \Big \{ \lambda_0\mu_0 +\sum_{j=1}^{d^{(m)}}\sum_{i=1}^n 
\tau_{j, v, g}^{(m)}\eta_{i, v, k}X_{i, j}^{(m)} \Big \} \\
 \gamma_{0, v, g, k}&=& \gamma_0 + \sum_{j=1}^{d^{(m)}}\sum_{i=1}^n \tau_{j, v, g}^{(m)}\eta_{i, v, k}\\
  \sigma_{0, v, g, k}^{2}&=&\frac{1}{\gamma_{0, v, g, k}}
 \Big \{\gamma_0\sigma_0^{2} + \lambda_0 \mu_0^2 \\
\no &+&\sum_{j=1}^{d^{(m)}}\sum_{i=1}^n \tau_{j, v, g}^{(m)}\eta_{i, v, k}({X_{i, j}^{(m)}})^2 - \lambda_{0, v, g, k} \mu_{0, v, g, k}^2 \Big \}.\\
\label{gaussupdate}
\end{eqnarray*}
Finally, the expectation of the conditional log-likelihood $\mathbb{E}_{q(\bb{\theta})}\Big [\log p(X_{i, j}^{(m)} | 
\bb{\theta}_{v, g, k}^{(m)})\Big ]$ is given by
\begin{eqnarray*}
&& -\frac{1}{2}
\Big \{ \frac{(X_{i, j}^{(m)}-\mu_{0, v, g, k})^2}{\sigma_{0, v, g, k}^2}+\frac{1}{\lambda_{0, v, g, k}}
+\log \sigma_{0, v, g, k}^2 \\
&&~~~~~~~~~+ \log (\gamma_{0, v, g, k}/2) -\psi(\gamma_{0, v, g, k}/2)+\log (2\pi) \Big \}.
\end{eqnarray*}

\subsection{Poisson distribution}
We denote Poisson distribution as $\mbox{Poisson}(\cdot|\lambda )$ where $\lambda$ is a rate parameter. 
The conjugate prior for $\lambda$ is given by
\begin{eqnarray*}
\lambda_{v, g, k} &\sim& \mbox{Ga}(\cdot|\alpha_0, \beta_0 ),
\end{eqnarray*}
where we set $\alpha_0$ and $\beta_0$ to one. 
It can be shown that the variational approximation is given by
\begin{eqnarray*}
 q_{\bb{\theta}^{(m)}}(\bb{\theta}^{(m)})=\prod_{v=1}^V\prod_{g=1}^G\prod_{k=1}^K 
 \mbox{Ga}(\lambda_{v, g, k}| \alpha_{0, v, g, k},\beta_{0, v, g, k}),
\end{eqnarray*}
where the hyperparameters are updated by
\begin{eqnarray*}
  \alpha_{0, v, g, k}&=&\alpha_0 + \sum_{j=1}^{d^{(m)}}\sum_{i=1}^n
 \tau_{j, v, g}^{(m)}\eta_{i, v, k}X_{i, j}^{(m)}\\
  \beta_{0, v, g, k}&=&\beta_0 +  \sum_{j=1}^{d^{(m)}}\sum_{i=1}^n
 \tau_{j, v, g}^{(m)}\eta_{i, v, k}.
 \label{poissonupdate}
\end{eqnarray*}
The expectation of the conditional log-likelihood becomes
\begin{eqnarray*}
&& X_{i, j}^{(m)}\{ \psi(\alpha_{0, v, g, k})-\psi(\beta_{0, v, g, k}) \}\\
&& ~~~~~~-\frac{\alpha_{0, v, g, k}}{\beta_{0, v, g, k}}-
\sum_{t=1}^{X_{i, j}^{(m)}}\log t.
\end{eqnarray*}

\subsection{Categorical/multinomial distribution}
 For a categorical feature $x$ ($x\in \{c_1, \ldots, c_H\}$), we denote categorical distribution as $\mbox{Cat}(\cdot|\bb{p})$ where 
$H$ is the number of categories, and 
$\bb{p}=(p_1, \ldots, p_H)$ are probabilities for each category with $\sum_{h=1}^H p_h=1$. We assume  the conjugate prior for $(p_1, \ldots, p_H)$,
\begin{eqnarray*}
(p_1, \ldots, p_H) &\sim & \mbox{Dirichlet}(\cdot | \bb{\rho}_0),
\end{eqnarray*}
where $\mbox{Dirichlet}(\cdot |\bb{\rho}_0)$ denotes a Dirichlet distribution with prior sample size $\bb{\rho}_0$. We set $\bb{\rho}_0$ to $(1, \ldots, 1)$. It can be shown that
\begin{eqnarray*}
 q_{\bb{\theta}^{(m)}}(\bb{\theta}^{(m)})=\prod_{v=1}^V\prod_{g=1}^G\prod_{k=1}^K 
 \mbox{Dirichlet}(\bb{p}_{v, g, k}| \bb{\rho}_{0, v, g, k}),
\end{eqnarray*}
where the hyperparameters are updated by
\begin{eqnarray*}
  \rho_{0, v, g, k, h} = \rho_{0, h} + \sum_{j=1}^{d^{(m)}}\sum_{i=1}^n 
 \tau_{j, v, g}^{(m)}\eta_{i, v, k}\mathbb{I}(X_{i, j}^{(m)}=c_h),
 \label{categoryupdate}
\end{eqnarray*}
where  $\rho_{0, v, g, k, h}$ denotes the $h$th element of $\bb{\rho}_{0, v, g, k}$.
The expectation of the log-likelihood is then given by
\begin{eqnarray*}
&& \sum_{h=1}^H \mathbb{I}(X_{i, j}^{(m)}=c_h)\{ \psi (\rho_{0, h, v, g, k})- 
\psi (\sum_{h'=1}^H \rho_{0, h', v, g, k}) \}.
\end{eqnarray*}
Since the categorical distribution differs depending on the number of categories $H$, we need to define different types of categorical distribution. 
Alternatively, for the purpose of simplicity, we can set $H$ to the maximum number of categories for different categorical features, and fit a single family of categorical distribution to all these features. 

More generally, in the case of multinomial distribution, the update equation and the expectation of
the log-likelihood becomes
\begin{eqnarray*}
&&  \rho_{0, v, g, k, h} = \rho_{0, h} + \sum_{j=1}^{d^{(m)}}\sum_{i=1}^n
 \tau^{(m)}\eta_{i, v, k}n_{i, j, h} \label{multinomialupdate} \\
 && \sum_{h=1}^H n_{i, j, h}\{ \psi (\rho_{0, h, v, g, k})- 
\psi (\sum_{h'=1}^H \rho_{0, h', v, g, k}) \} \\
\no && ~~~~~~~~~~~~~~~~~~~~~~~~~~~~~~~~+\log 
\begin{pmatrix}
\sum_{h=1}^H n_{i, j, h}\\
n_{i, j, 1}, \ldots,  n_{i, j, H}
\end{pmatrix}
,
\end{eqnarray*}
where $n_{i, j, h}$ is the number of category $c_h$ in $X_{i, j}^{(m)}$; the last term is the logarithm of 
multinomial coefficients. 

\newpage
\section{List of features for clinical data} \label{listfeatures}
\begin{table}[!h]
\footnotesize
\caption{\it List of features for clinical data}
\label{association}
\begin{tabular}{l}
\multicolumn{1}{l}{Features}\\
\hline
  \textbf{Numerical features}\\
  ~~~BAS (Behavioral Activation Scale),\\
  ~~~BDNF (Quantity of brain-derived \\
  ~~~~~~~~~~~~~~~~neurotrophic factor in blood),\\
  ~~~BDI2 (Beck Depression Inventory), \\
  ~~~BIS (Behavioral Inhibition Scale),\\
  ~~~CATS (Child Abuse and Trauma Scale),\\
  ~~~Cortisol (Quantity of cortisol in blood),\\
  ~~~GAF (Global Assessment of Functioning),\\
  ~~~PHQ9 (Patient Health Questionnaire), \\
  ~~~HRSD17 (Hamilton Rating Scale for Depression),\\
  ~~~JART (Adult reading test),   \\
  ~~~LES (Life Experiences Survey),\\
  ~~~PANASP (Positive Affect Schedule),\\
  ~~~PANASN (Negative Affect Schedule),\\
  ~~~SHAPS (Snaith-Hamilton Pleasure Scale),\\
  ~~~STAI (State-Trait Anxiety Inventory),\\ 
  ~~~N, E, O, A, C \\
 ~~~(Five factors in revised NEO Personality Inventory) \\
 \textbf{Categorical features} \\
 ~~~Sex \\
 ~~~\textit {miniA-P (Mini-International Neuropsychiatric Interview)},\\
 ~~~\textit{A-P corresponds to the following psychiatric symptoms}: \\
 ~~~Major depressive disorder (A),\\
 ~~~Dysthymia (B), Suicide risk (C),\\
 ~~~Mania (D), Panic disorder (E), Agoraphobia (F), \\ 
 ~~~Social phobia (G),\\
 ~~~Obsessive compulsive disorder (H),\\
 ~~~PTSD (I), Alcohol dependence and abuse (J), \\
 ~~~Drug dependence and abuse (K), \\
 ~~~Psychotic disorder (L), Anorexia (M), Bulimia (N),\\
 ~~~Generalized anxiety disorder (O),\\
 ~~~Antisocial personality disorder (P), \\
 ~~~\textit {SNPs 1-8: Single Nucleotide Polymorphisms that} \\
 ~~~ \textit{are located in the following genome sites, respectively.}\\
 ~~~ \textit{(in parenthesis are the relevant gene functions)} \\
~~~ rs1187323 (NTRK2), rs34118353 (5HT1a receptor), \\
~~~ rs3756318 (NTRK2), rs3813929 (5HT2c receptor), \\
~~~  rs45554739 (NTRK2), rs56384968 (SLC6A4), \\
~~~ rs6265 (BDNF), rs6294 (5HT1a receptor) \\
\hline
\end{tabular}
\end{table}
\end{appendices}


\input{PaperTheory8.bbl}
\bibliographystyle{plain}

\end{document}

%% file: clustermatMulface.txt
\begin{small}\begin{tabular}{|l|c|c|c|c|}
\multicolumn{5}{c}{(a) Mul} \\
\hline
&\textbf{T1}&\textbf{T2}&\textbf{T3}&\textbf{T4}\\\hline
\textbf{C1}&8&0&0&2\\\hline
\textbf{C2}&0&8&0&0\\\hline
\textbf{C3}&0&0&8&0\\\hline
\textbf{C4}&0&0&0&6\\\hline
\end{tabular}
\end{small}

%% file: clustermatCOALAface.txt
\begin{small}\begin{tabular}{|c|c|c|c|c|}
\multicolumn{5}{c}{(b) COALA} \\
\hline
&\textbf{T1}&\textbf{T2}&\textbf{T3}&\textbf{T4}\\\hline
\textbf{C1}&2&0&6&1\\\hline
\textbf{C2}&6&0&1&7\\\hline
\textbf{C3}&0&5&1&0\\\hline
\textbf{C4}&0&3&0&0\\\hline
\end{tabular}
\end{small}

%% file: clustermatDecKface.txt
\begin{small}\begin{tabular}{|l|c|c|c|c|}
\multicolumn{5}{c}{(c) DecK} \\
\hline
&\textbf{T1}&\textbf{T2}&\textbf{T3}&\textbf{T4}\\\hline
\textbf{C1}&0&0&2&2\\\hline
\textbf{C2}&2&3&1&2\\\hline
\textbf{C3}&4&4&4&3\\\hline
\textbf{C4}&2&1&1&1\\\hline
\end{tabular}
\end{small}

%% file: clustermatRestface.txt
\begin{small}\begin{tabular}{|l|c|c|c|c|}
\multicolumn{5}{c}{(d) rMul} \\
\hline
&\textbf{T1}&\textbf{T2}&\textbf{T3}&\textbf{T4}\\\hline
\textbf{C1}&0&8&0&0\\\hline
\textbf{C2}&0&0&8&0\\\hline
\textbf{C3}&1&0&0&5\\\hline
\textbf{C4}&4&0&0&2\\\hline
\textbf{C5}&3&0&0&1\\\hline
\end{tabular}
\end{small}

%% file: clustermatMulface64useid.txt
\begin{small}\begin{tabular}{|l|c|c|}
\multicolumn{3}{c}{(a) Mul} \\
\hline
&\textbf{T1}&\textbf{T2}\\\hline
\textbf{C1}&32&0\\\hline
\textbf{C2}&0&25\\\hline
\textbf{C3}&0&7\\\hline
\multicolumn{3}{c}{  } \\
\end{tabular}
\end{small}

%% file: clustermatCOALAface64useid.txt
\begin{small}\begin{tabular}{|l|c|c|}
\multicolumn{3}{c}{(b) COALA} \\
\hline
&\textbf{T1}&\textbf{T2}\\\hline
\textbf{C1}&32&0\\\hline
\textbf{C2}&0&32\\\hline
\multicolumn{3}{c}{} \\
\multicolumn{3}{c}{} \\
\end{tabular}
\end{small}

%% file: clustermatDecKface64useid.txt
\begin{small}\begin{tabular}{|l|c|c|}
\multicolumn{3}{c}{(c) DecK} \\
\hline
&\textbf{T1}&\textbf{T2}\\\hline
\textbf{C1}&3&32\\\hline
\textbf{C2}&29&0\\\hline
\multicolumn{3}{c}{} \\
\multicolumn{3}{c}{} \\
\end{tabular}
\end{small}

%% file: clustermatRestface64useid.txt
\begin{small}\begin{tabular}{|l|c|c|}
\multicolumn{3}{c}{(d) rMul} \\
\hline
&\textbf{T1}&\textbf{T2}\\\hline
\textbf{C1}&32&0\\\hline
\textbf{C2}&0&15\\\hline
\textbf{C3}&0&13\\\hline
\textbf{C4}&0&4\\\hline
\end{tabular}
\end{small}

%% file: clustermatMulface64.txt
\begin{small}\begin{tabular}{|l|c|c|c|c|}
\multicolumn{5}{c}{(a) Mul} \\
\hline
&\textbf{T1}&\textbf{T2}&\textbf{T3}&\textbf{T4}\\\hline
\textbf{C1}&7&8&0&8\\\hline
\textbf{C2}&1&0&0&8\\\hline
\textbf{C3}&0&1&7&0\\\hline
\textbf{C4}&0&7&1&0\\\hline
\textbf{C5}&7&0&0&0\\\hline
\textbf{C6}&0&0&5&0\\\hline
\textbf{C7}&1&0&3&0\\\hline
\end{tabular}
\end{small}

%% file: clustermatCOALAface64.txt
\begin{small}\begin{tabular}{|l|c|c|c|c|}
\multicolumn{5}{c}{(b) COALA} \\
\hline
&\textbf{T1}&\textbf{T2}&\textbf{T3}&\textbf{T4}\\\hline
\textbf{C1}&8&1&1&6\\\hline
\textbf{C2}&0&7&0&0\\\hline
\textbf{C3}&0&0&9&0\\\hline
\textbf{C4}&8&8&6&10\\\hline
\multicolumn{5}{c}{} \\
\multicolumn{5}{c}{} \\
\multicolumn{5}{c}{} 
\end{tabular}
\end{small}

%% file: clustermatDecKface64.txt
\begin{small}\begin{tabular}{|l|c|c|c|c|}
\multicolumn{5}{c}{(c) DecK} \\
\hline
&\textbf{T1}&\textbf{T2}&\textbf{T3}&\textbf{T4}\\\hline
\textbf{C1}&7&4&0&2\\\hline
\textbf{C2}&0&3&14&3\\\hline
\textbf{C3}&6&3&0&10\\\hline
\textbf{C4}&3&6&2&1\\\hline
\multicolumn{5}{c}{} \\
\multicolumn{5}{c}{} \\
\multicolumn{5}{c}{} \\
\end{tabular}
\end{small}

%% file: clustermatRestface64.txt
\begin{small}\begin{tabular}{|l|c|c|c|c|}
\multicolumn{5}{c}{(d) rMul} \\
\hline
&\textbf{T1}&\textbf{T2}&\textbf{T3}&\textbf{T4}\\\hline
\textbf{C1}&5&3&2&4\\\hline
\textbf{C2}&5&0&8&0\\\hline
\textbf{C3}&0&0&6&4\\\hline
\textbf{C4}&1&1&0&8\\\hline
\textbf{C5}&2&7&0&0\\\hline
\textbf{C6}&3&5&0&0\\\hline
\end{tabular}
\end{small}

%% file: Matrix1.tex
\begin{small}
\begin{tabular}{|l|c|c|c|}
\multicolumn{4}{c}{(a) Mul} \\
\hline
&\textbf{T1}&\textbf{T2}&\textbf{T3}\\\hline
\textbf{C1}&0&14&6\\\hline
\textbf{C2}&14&0&0\\\hline
\textbf{C3}&1&1&5\\\hline
\textbf{C4}&0&0&2\\\hline
\end{tabular}
\end{small}

%% file: MatrixCOALA2.tex
\begin{small}\begin{tabular}{|l|c|c|c|}
\multicolumn{4}{c}{(b) COALA} \\
\hline
&\textbf{T1}&\textbf{T2}&\textbf{T3}\\\hline
\textbf{C1}&15&15&10\\\hline
\textbf{C2}&0&0&2\\\hline
\textbf{C3}&0&0&1\\\hline
\multicolumn{4}{c}{} \\
\end{tabular}
\end{small}

%% file: MatrixK1.tex
\begin{small}\begin{tabular}{|l|c|c|c|}
\multicolumn{4}{c}{(c) DecK} \\
\hline
&\textbf{T1}&\textbf{T2}&\textbf{T3}\\\hline
\textbf{C1}&14&0&0\\\hline
\textbf{C2}&0&0&1\\\hline
\textbf{C3}&1&15&12\\\hline
\multicolumn{4}{c}{} \\
\end{tabular}
\end{small}

%% file: MatrixRest6.tex
\begin{small}\begin{tabular}{|l|c|c|c|}
\multicolumn{4}{c}{(d) rMul} \\
\hline
&\textbf{T1}&\textbf{T2}&\textbf{T3}\\\hline
\textbf{C1}&14&0&2\\\hline
\textbf{C2}&1&5&4\\\hline
\textbf{C3}&0&3&6\\\hline
\textbf{C4}&0&7&1\\\hline
\end{tabular}
\end{small}